\documentclass[10pt,twocolumn,letterpaper]{article}

\usepackage[pagenumbers]{cvpr} 

\usepackage{threeparttable}
\definecolor{orangehl}{HTML}{EBAF9B}
\definecolor{rose}{HTML}{FF7C80}
\definecolor{mintgreen}{HTML}{99FF99}
\definecolor{cvprblue}{rgb}{0.21,0.49,0.74}
\usepackage[pagebackref,breaklinks,colorlinks,allcolors=cvprblue]{hyperref}
\usepackage{algpseudocode}
\usepackage[ruled,vlined]{algorithm2e}

\title{\includegraphics[width=0.5cm]{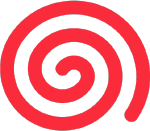}verthinking Causes Hallucination:\\ Tracing Confounder Propagation in Vision Language Models}
\author{
Abin Shoby$^{1\dagger}$ \qquad 
Ta Duc Huy$^{1\dagger}$\\
Tuan Dung Nguyen $^{2}$  \qquad Minh Khoi Ho$^{2,3}$ \qquad Qi Chen$^{1}$\\
Anton van den Hengel$^{1}$
\qquad Phi Le Nguyen$^{2}$
\qquad Johan W. Verjans$^{1}$
\\ Vu Minh Hieu Phan$^{1}$\thanks{Lead author.}
\\
\small{$^{1}$ Australian Institute for Machine Learning, University of Adelaide}
\\
\small{$^{2}$ Hanoi University of Science and Technology}  \qquad
\small{$^{3}$ Mohamed bin Zayed University of Artificial Intelligence} \\
\small{$^{\dagger}$ Equal contribution}
}

\begin{document}
\maketitle
\begin{abstract}

Vision Language models (VLMs) often hallucinate non-existent objects. Detecting hallucination is analogous to detecting deception: a single final statement is insufficient, one must examine the underlying reasoning process. Yet existing detectors rely mostly on final-layer signals. Attention-based methods assume hallucinated tokens exhibit low attention, while entropy-based ones use final-step uncertainty. Our analysis reveals the opposite: hallucinated objects can exhibit peaked attention due to contextual priors; and models often express high confidence because intermediate layers have already converged to an incorrect hypothesis.
We show that the key to hallucination detection lies within the \textbf{model's thought process}, not its final output. By probing decoder layers, we uncover a previously overlooked behavior, overthinking: models repeatedly revise object hypotheses across layers before committing to an incorrect answer. Once the model latches onto a \textup{confounded} hypothesis, it can propagate through subsequent layers, ultimately causing hallucination. To capture this behavior, we introduce the \textbf{Overthinking Score}, a metric to measure how many competing hypotheses the model entertains and how unstable these hypotheses are across layers. This score significantly improves hallucination detection: 78.9\% F1 on MSCOCO and 71.58\% on AMBER.
\end{abstract}

\section{Introduction}
\label{sec:intro}

\begin{figure}
    \centering
    \includegraphics[width=1\linewidth]{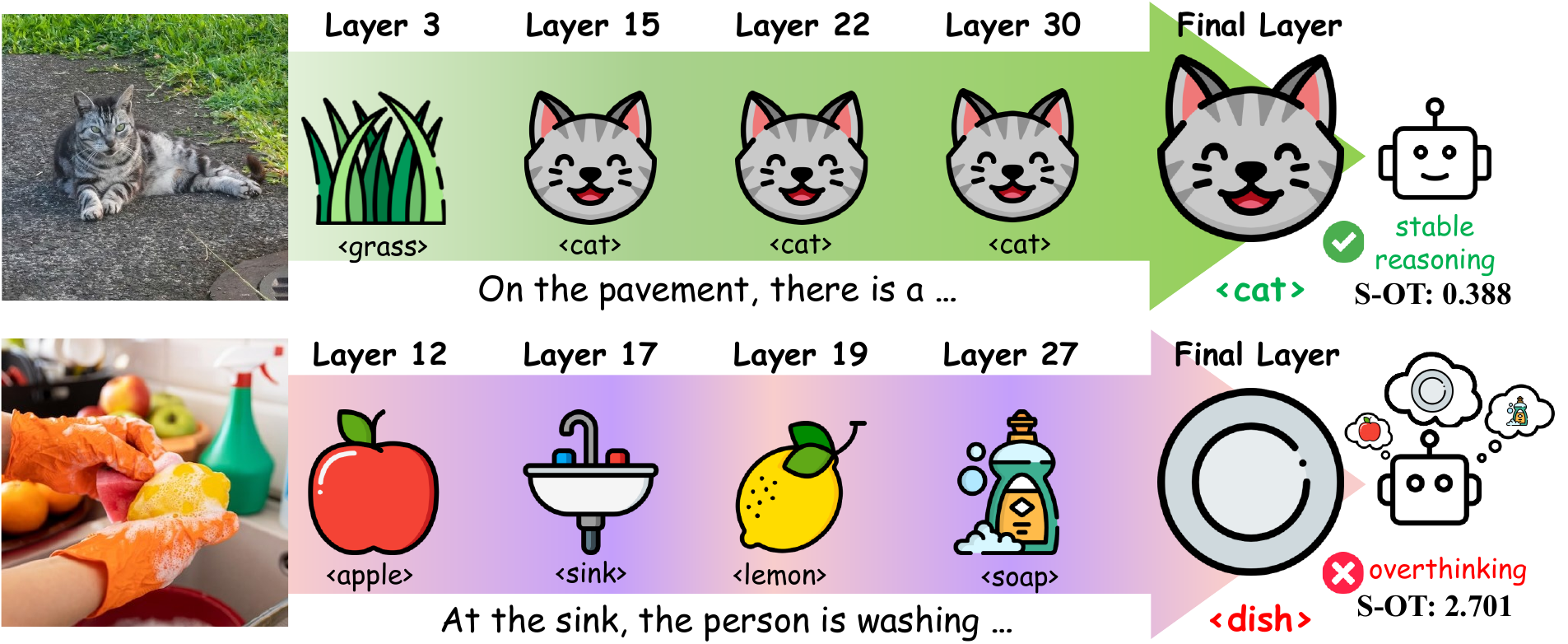}
    \caption{\textbf{Overthinking leads to object hallucination in Vision-Language Models.} Each column corresponds to the top predicted object token at a decoder layer, illustrating the model’s layer-wise reasoning progression. We propose Overthinking Score (S-OT) to measure how much the model shifts among objects across layers.
    \textit{Top}: the model demonstrates \textit{stable reasoning}, quickly converges on a consistent concept (cat) across decoder layers, yielding low S-OT.
    \textit{Bottom}: the model shows \textit{overthinking}, hesitating between semantically co-occurring objects or ``confounders" (sink, soap) that bias it towards confidently producing a hallucinated answer (dish) captured by high S-OT.
    }
    \label{fig:teaser}
\end{figure}



Hallucination is a persistent problem in Vision Language models (VLMs), where the model describes objects that are not present in the input image.
Early hallucination detection approaches introduce external judge models~\cite{woodpecker, unihd, haelm, faithscore} to verify generated responses using auxiliary evaluators. While effective in some cases, these methods are limited by their dependence on additional large models, which makes them computationally expensive and often unreliable when the judge itself inherits similar biases from the base model.

Attention-based methods~\cite{svar,damro,opera,metatoken} have been proposed to detect hallucinated objects using visual attention magnitudes. These approaches rely on the assumption that real objects exhibit 
high attention on real objects.
However, our correlation analysis challenges this premise: we find that hallucinated objects can 
exhibit high attention magnitude on spurious region (Fig.~\ref{fig:attention_fail}),
showing that attention intensity does not reliably indicate hallucination.


\begin{figure*}
    \centering
    \includegraphics[width=\linewidth]{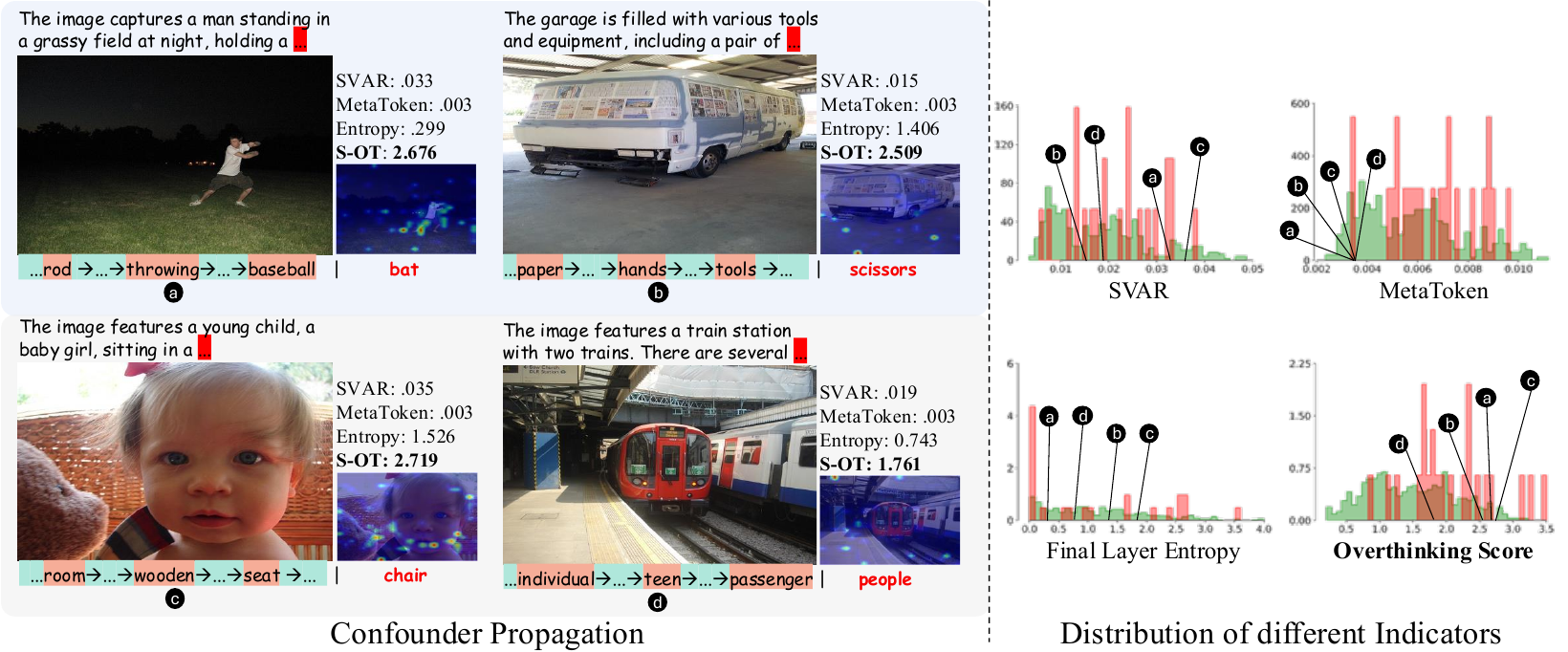}
    \caption{
\textit{Left:} \textbf{Confounder Propagation examples.} In each example, we illustrate the image and its prefix prompt on top. Under the image,  
we highlight  \colorbox{orangehl}{confounders}
in the intermediate layers prior to the  \textcolor{red}{hallucinated final layer token}. We also list the values of different hallucination indicators: SVAR~\cite{svar}, MetaToken~\cite{metatoken}, Final Layer Entropy, and our proposed \textbf{Overthinking Score (S-OT)}. The attention map of the fifth layer is also presented following~\cite{svar}.
Existing methods ignore intermediate-layer token dynamics, missing hallucinations. In contrast, our approach detects by tracing confounder propagation.
\textit{Right:} Histograms of SVAR, MetaToken, Final-Layer Entropy, and our \textbf{Overthinking Score} are shown, with the example positions highlighted. Existing indicators show high overlapping between \textcolor[rgb]{0.6,0.8,0.6}{Real}  and \textcolor[rgb]{1.0,0.514,0.514}{Hallucinated} object distribution, while our Overthinking Score offers noticeably better separation.
}
    \label{fig:confounders}
\end{figure*}


Recent approaches attempt to diagnose hallucination by measuring entropy or uncertainty at the final decoding layer~\cite{metatoken,halloc,vl_uncertainty}, assuming hallucination corresponds to output ambiguity. Interestingly, we find that hallucination can arise in earlier stages. 
Intermediate layers can activate several object hypotheses, which may include a \textit{confounder}.
Here, the confounders are plausible concepts that bias the model to hallucinate the object in the final prediction.
For example, as shown in Fig.~\ref{fig:teaser}, the confounding concepts ``sink" and ``soap" emerge in the intermediate layers, influencing the VLM to generate ``dish" at the final layer, which is nonexistent but semantically aligned with sink and soap.
Hence, VLMs can hallucinate 
confounded objects that are consistent with the scene context rather than the present objects in the image. We illustrate this effect in Fig.~\ref{fig:confounders},
where the model hallucinates at the final decoding layer after multiple confounders emerge in intermediate layers, a process we term ``confounder propagation". This phenomenon reflects a well-known issue in deep networks: confounders. 
Prior works have shown that VLMs are prone to encode confounding features such as background texture or co-occurrence objects~\cite{rope,lu2025mitigating,zhou2023analyzing}.


Mechanistically, the final decoder layer is conditioned on its preceding layers' representation. Using LogitLens~\cite{logitlens}, we decode intermediate-layer token features into vocabulary and find that they exhibit strong semantic alignment with the final-layer token (Tab.~\ref{tab:influence_of_intermediate_layer_to_final_layer}).
Hence, these intermediate-layer objects can semantically influence the formation of the final-layer object. Empirically, we observe that: when intermediate layers yield 
excessive distinct
object tokens,
the model is prone to hallucination. We hypothesize that when the model considers many alternative object hypotheses, it is more likely to include a confounder and increases the likelihood of predicting a nonexistent object.
Our analysis shows a positive correlation between the number of objects and confounder propagation rate as illustrated in Fig.~\ref{fig:corr_conf_rates}-\textit{Right}.
Tracking the proliferation of these latent object hypotheses throughout the layers reveals the confounder propagation phenomenon that final layer analysis alone fails to capture.
We also find that hallucinations appearing in intermediate layers often persist to the final output. Intermediate layers with higher entropy tend to produce more hallucinations at the final layer as shown in Fig.~\ref{fig:corr_conf_rates}-\textit{Left}.
In other words, when uncertainty accumulates throughout the layers, the model is more likely to hallucinate.

This behavior reflects \textit{overthinking} (Fig.~\ref{fig:teaser}), where the model forms multiple object hypotheses before reaching a final answer, increasing the chance of confounders that facilitate hallucination. This motivates us to propose the \textbf{Overthinking Score} to quantify: (1) the number of unique top-1 tokens; and (2) the uncertainty across layers. Intuitively, if the model considers too many objects and exhibits high uncertainty in the intermediate layers, it is more likely to hallucinate. Unlike prior methods that rely solely on the final layer, the Overthinking Score aggregates all layers to explicitly capture \textit{confounder propagation} for robust hallucination detection.
We summarize our contributions:
\begin{enumerate}
    \item We conduct a layer-wise investigation of VLMs and uncover the \textit{confounder propagation} phenomenon that drives hallucination.
    \item We propose the \textit{Overthinking Score} that effectively captures this phenomenon and achieves strong predictive performance.
    \item Extensive experiments across multiple benchmarks show that our overthinking-based detector consistently surpasses prior hallucination detection methods.
\end{enumerate}



\section{Related Work}

\textbf{Hallucination Analysis.}
Several works examine internal behaviors that give rise to hallucination in LVLMs. SVAR~\cite{svar} studies layer-wise attention and introduces the Visual Attention Ratio (VAR), showing that real object tokens exhibit higher mid-layer attention to image patches. Yang et al.~\cite{yang2024tokenlevel} analyze token-level attributions using integrated gradients and attention rollout, finding that hallucinated tokens attend less to relevant patches and more to language-prior positions. DAMRO~\cite{damro} traces attention-head divergence patterns correlated with hallucinations. While these studies reveal useful attention signatures, they focus primarily on attention magnitude and do not track how intermediate token hypotheses semantically shape the final prediction. As a result, they overlook the layer-wise propagation of confounding object representations, which we show to be a major driver of hallucination.\\
\textbf{Hallucination Detection.}
Early hallucination detection methods relies on grounding via auxiliary models~\cite{woodpecker, unihd, haelm, faithscore}.
Uncertainty-based detectors adapted from LLMs (e.g., SelfCheckGPT~\cite{manakul-etal-2023-selfcheckgpt}, Semantic Entropy~\cite{kuhn2023semantic}) estimate prediction ambiguity but operate exclusively at the final layer. Recent token-level approaches such as MetaToken~\cite{metatoken}, SVAR~\cite{svar}, and HalLoc~\cite{halloc} extract handcrafted features or attention statistics, or train multimodal classifiers with token-level labels. However, these methods largely rely on surface signals: final-layer entropy, attention magnitudes, or feature heuristics, and do not model how confounding activations accumulate and persist across layers. Our work fills this gap by formalizing confounder propagation and proposing the Overthinking Score, which captures both semantic branching and layer-wise uncertainty dynamics that current methods overlook.
\section{Methodology}
We first characterize \textit{confounder propagation} through a layer-wise analysis of VLMs (Sec.~\ref{sec:problem_analysis}), then leverage these insights to design the \textit{Overthinking Score} for our hallucination detection pipeline (Sec.~\ref{sec:hallu_pipeline}).

\label{sec:method}
\subsection{Problem Analysis}
\label{sec:problem_analysis}
We focus on white-box approaches that inspect model internals rather than black-box judge systems dependent on external evaluators~\cite{woodpecker, unihd, haelm, faithscore}. Our goal is to identify internal indicators of hallucination.
We find that hallucinations often occur in scenes with strong contextual priors.
Current methods based on attention or final-layer uncertainty cannot capture \textit{confounder propagation}, the internal accumulation of contextual bias that leads to hallucination at the final layer. We propose three hypotheses to formalize this layer-wise phenomenon.

\textbf{H1.}\textit{ Strong contextual priors cause hallucinated object tokens to exhibit visual attention magnitude that is comparable to real objects.}

\noindent 
Attention-based methods~\cite{svar,damro,opera,metatoken} assume that real objects attract stronger visual attention than hallucinated ones. However, in scenes with strong contextual priors (e.g., kitchen-microwave), hallucinated objects often receive attention comparable to, or even higher than, real ones.
To examine the impact of contextual priors, we employ~\cite{sentence_transformer} to compute the semantic correlation between the tokens referring to  the scene versus each generated object to estimate prior strength, and select highly correlated samples as \textit{strong-prior} cases. We then compare the distribution of visual-attention indicators used by SVAR~\cite{svar} and MetaToken~\cite{metatoken} between real and hallucinated objects. As shown in Fig.~\ref{fig:confounders}-\textit{Right}
, their distributions largely overlap, which supports the hypothesis, suggesting that visual attention magnitude cannot reliably distinguish hallucination under a strong contextual prior.

\textbf{H2.}\textit{ Confounders emerging in intermediate layers propagate to induce hallucination, a phenomenon that final-layer uncertainty cannot capture.}

\noindent We first examine whether intermediate layers contextually influence the formation of the final layer prediction. To test this, we evaluate the semantic alignment between the top-1 tokens decoded from each intermediate layer and the final layer token across multiple VLMs. We use LogitLens~\cite{logitlens} to decode intermediate activations into the vocabulary space and~\cite{sentence_transformer} to compute the semantic alignment. The consistently high alignment observed (Tab.~\ref{tab:influence_of_intermediate_layer_to_final_layer})
implies a semantic influence of intermediate tokens on the final decoded object.

\begin{table}[]
    \centering
    \begin{tabular}{cccc}
    \toprule
        & \textbf{LLaVA-1.5} & \textbf{Gemma-3} & \textbf{Qwen3-VL} \\
        \midrule
        Hallu. & 40.6\% & 47.9\% & 58.6\% \\
        Real  & 41.6\% & 42.4\% & 55.1\% \\
        All & 41.5\% & 43.2\% & 55.5\% \\
        \bottomrule
    \end{tabular}
    \caption{Semantic alignment between intermediate-layer tokens and the final-layer decoded token for LLaVA-1.5, Gemma-3, and Qwen3-VL. We report the mean alignment for hallucinated tokens, real tokens, and all tokens.
    The high alignment implies a semantic influence from the intermediate layers to the final layer, indicating \textit{confounder propagation}.}
    \label{tab:influence_of_intermediate_layer_to_final_layer}
\end{table}
\noindent Next, we define ``confounder propagation" as the specific cases where objects emerging in intermediate layers influence the final layer to generate contextually correlated but \textit{nonexistent} object (Fig.~\ref{fig:confounders}-\textit{Left}).
A sample is considered to exhibit confounder propagation when \textit{the final layer hallucinated token shows strong semantic correlation with tokens decoded from intermediate layers}.
Our analysis shows that confounder propagation is significant,
 accounting for \textbf{63.69\%} of hallucinations in LLaVA-1.5, \textbf{82.73\%} in Gemma-3, and \textbf{85.46\%} in Qwen3-VL on MS-COCO
(see Supp.). 

\begin{figure}
    \centering
    \includegraphics[width=1\linewidth]{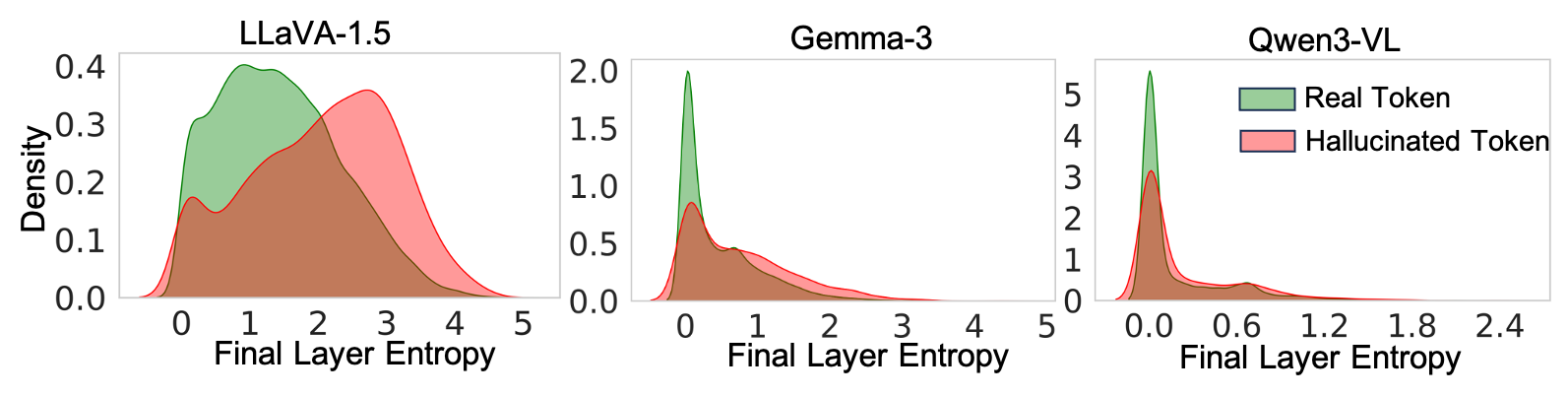}
    \caption{Distribution of final layer entropy in LLaVA-1.5, Gemma-3 and Qwen3-VL for hallucinated and real tokens.The strong overlap shows the weakness of entropy as a hallucination predictor.}
    \label{fig:exp4b}
\end{figure}

Because these confounders bias the representation before decoding, the model often produces confident (low-entropy) predictions even when hallucinating. Empirically, we show that the entropy distributions of hallucinated and real objects overlap substantially (Fig.~\ref{fig:exp4b}),
 indicating that hallucination is not necessarily accompanied by uncertainty. These results demonstrate that while final layer entropy analysis provides some diagnostic value of hallucination, it 
 masks the underlying
 confounder propagation, which can only be captured through intermediate layer analysis.

\textbf{H3.}\textit{ More unique object hypotheses across layers increase the chance that a confounder emerges which facilitates hallucination.}

\noindent We observe that certain samples trigger a large number of unique object hypotheses across layers. This behavior increases the likelihood that a confounder will arise. Since intermediate layer representations can influence the final token as shown in \textbf{H2}, such confounder can propagate and bias  the final prediction to hallucinate.

\begin{figure}
    \centering
    \includegraphics[width=1\linewidth, height=2cm]{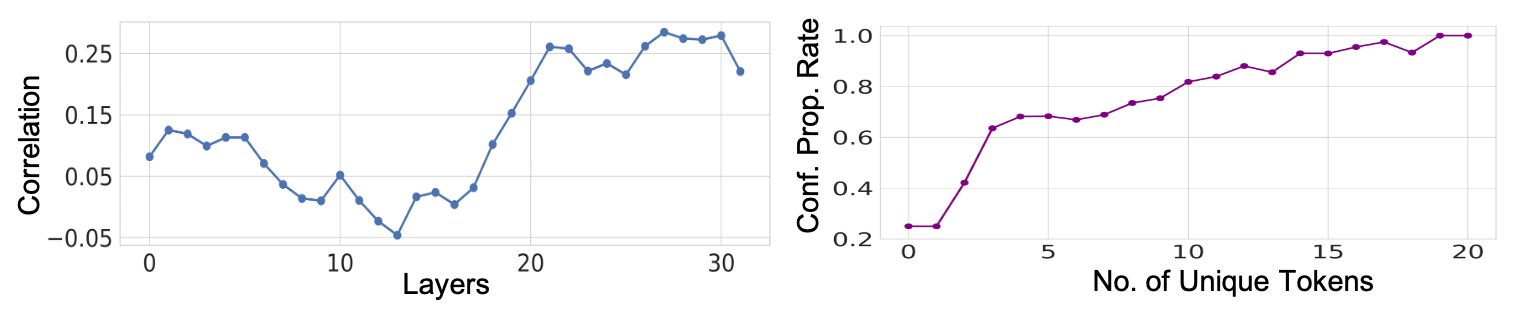}
    \caption{\textit{Left:} Correlation between per-layer average entropy and hallucination rate. Most layers exhibit positive correlation, suggesting that increased uncertainty across depth contributes to hallucination. \textit{Right:} Confounder propagation rate versus the number of unique tokens in the intermediate layers. A higher number of unique tokens is associated with an increased likelihood of confounder propagation. Both measured on LlaVA-1.5}
    \label{fig:corr_conf_rates}
\end{figure}

\noindent Empirically, we find a positive correlation between the number of unique intermediate tokens and the measured confounder-propagation rate  as shown in Fig.~\ref{fig:corr_conf_rates}-\textit{Right}.
This indicates that when the model reasons over too many object alternatives, it becomes more susceptible to confounder propagation.
Additionally, our analysis shows that intermediate layers with higher entropy tend to produce more hallucinations at the final layer (Fig.~\ref{fig:corr_conf_rates}-\textit{Left})
, indicating that uncertainty accumulation further interacts with confounder propagation to amplify hallucination.

\noindent\textbf{Discussion.} 
Visual attention based methods often fail under strong contextual priors, where hallucinated and real objects receive comparable attention magnitudes.
Furthermore, our analysis shows that when the model ``overthinks", which explores too many object hypotheses; and accumulates uncertainty across layers, it becomes more prone to hallucination.
To capture both effects, we introduce the \textit{Overthinking Score} in Sec.~\ref{sec:feat_extract}, which captures the internal layer dynamics and remains effective even in strong contextual prior scenes.

\subsection{Hallucination Detection Pipeline}
\label{sec:hallu_pipeline}
\begin{figure*}
    \centering
    \includegraphics[width=\linewidth]{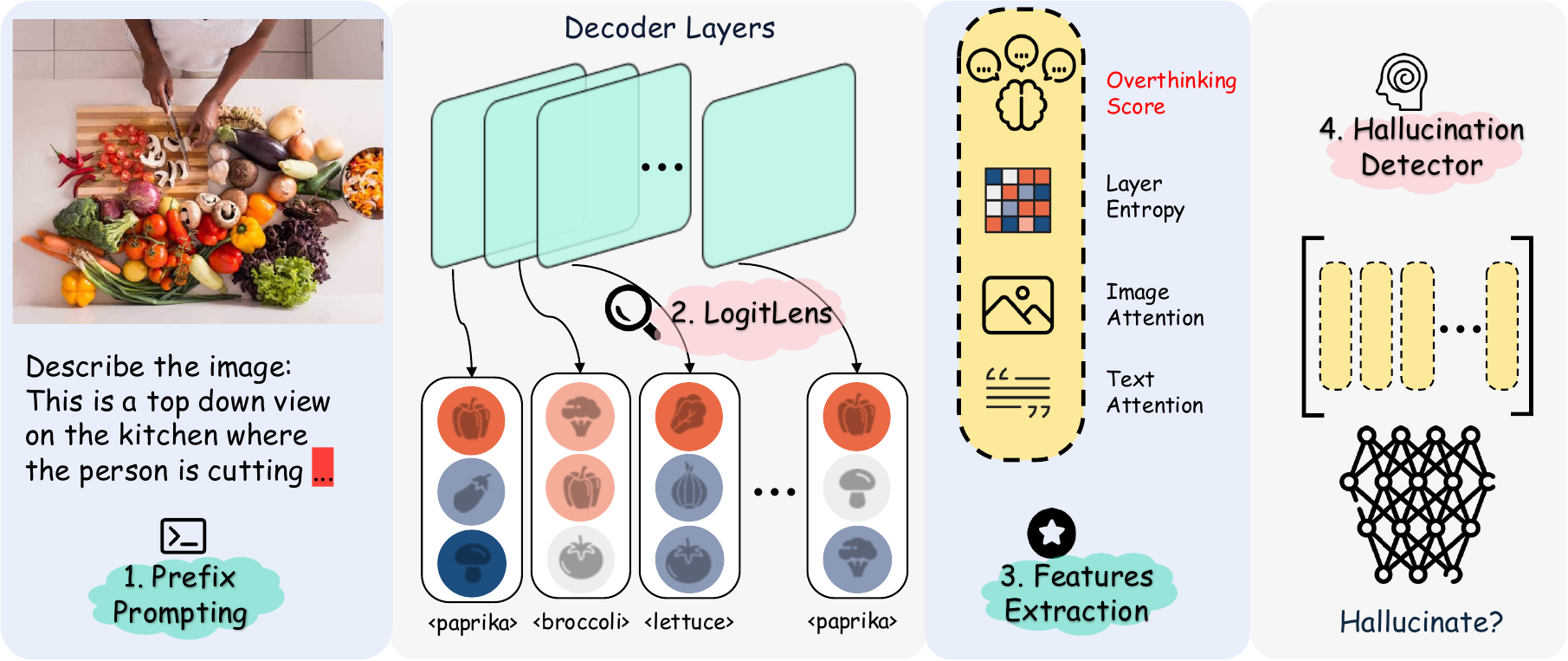}
    \caption{\textbf{Hallucination detection pipeline.}
1) We begin with \textit{Prefix Prompting}, where the model is asked to predict the next token given an image and a partial prompt.
2) We apply \textit{LogitLens} to extract the top-$p$ tokens at each decoder layer, revealing how the model’s intermediate hypotheses evolve.
3) In the \textit{Feature Extraction} step, the top-$p$ tokens distributions are used to compute the \textcolor{red}{Overthinking Score}  and layer-wise entropy, while image and text attention are measured with respect to the generated token. All features are concatenated into a single feature vector.
4) A \textit{Hallucination Detector} is trained on these features to identify hallucinated cases.}
    \label{fig:method}
\end{figure*}

We illustrate our Hallucination detection pipeline in Fig.~\ref{fig:method}.
Given a Vision Language Model, we provide it with an input image and a partial text prompt for the model to generate the next word in the sequence (Sec.~\ref{sec:prefix_prompt}).
The objective is to determine whether the predicted token refers to a real object in the image or a hallucinated one.
To this end, we analyze the internal token-level dynamics of the VLM during the generation of the token (Sec.~\ref{sec:logit_lens}) and compute an \textit{Overthinking Score}, which quantifies how frequently the model revises its beliefs across layers before committing to a final prediction.
Alongside this score, we extract complementary features including layer-wise entropy, image attention, and text attention with respect to the generated token.
These features are aggregated into a single feature vector (Sec.~\ref{sec:feat_extract}).
A labeled set of these feature vectors, corresponding to real and hallucinated objects, is used to train a lightweight Hallucination Detector (Sec.~\ref{sec:detection_model}).

\subsubsection{Prefix Prompting}
\label{sec:prefix_prompt}
Following~\cite{svar}, given a VLM with $L$ transformer layers and a token vocabulary ${v}^V$, we first prompt the model with a general instruction: \textit{``Describe this image."} The model then produces a detailed response containing multiple object mentions.
We then extract each mentioned object from the response and label it as either real or hallucinated, producing a set of pairs $(t, y_t)$, where $t$ denotes the token position of the object mention and $y_t$ is the hallucination label ($y_t = 1$ for hallucinated, $y_t = 0$ for real).

For each detected object, we construct a targeted prefix prompt using the portion of the model’s original response preceding that object token. The resulting prefix $x_{1:t-1}$, together with the image $I$, is fed back into the model to predict the next token $x_t$. The task is to determine whether the generated token $x_t$ is real or hallucinated.

The prefix is designed such that the expected next token should correspond to an object present in the image. 
For example, for an image depicting a kitchen scene, the prefix prompt may take the form: \textit{``In this image, I can see a ..."} encouraging the model to complete the sentence with a relevant object such as \textit{``plate"} or \textit{``sink"}. 
By structuring the prompt in this way, we constrain the decoding process to object-referential tokens.

\subsubsection{Tracing Internal Reasoning Dynamics}
\label{sec:logit_lens}
We trace the internal reasoning dynamics of the VLM by analyzing the top-$p$ tokens at each transformer layer $\ell$.
To access intermediate layer predictions, we employ LogitLens~\cite{logitlens}, a diagnostic technique that projects hidden representations from any transformer layer into the model output vocabulary space using the final linear head.
We denote $p_\ell(v)$ as the token distribution at layer $\ell$, which is computed as:
\begin{equation}
    p_\ell(v) = \text{softmax}(W \cdot \text{LayerNorm}(h_\ell)), \quad \ell = 1, \dots, L,
\end{equation}
where $h_\ell \in \mathbb{R}^{d}$ denotes the hidden state vector at layer $\ell$ for the current token, 
and $W \in \mathbb{R}^{V \times d}$ is the output projection matrix mapping from the hidden dimension $d$ to the vocabulary size $V$.
The resulting $p_\ell(v) \in \mathbb{R}^{V}$ represents the probability distribution over the vocabulary at the layer $\ell$.
By examining the top-$p$ tokens from $p_\ell(v)$ across all layers, we can observe how the model's dominant hypotheses evolve through the decoding process.
These layer-wise token trajectories indicate whether the model maintains a consistent belief or frequently revises it, a behavior that reflects overthinking.\\
\textbf{Discussion.} While PROJECTAWAY~\cite{projectaway} also leverages a logit-lens strategy to inspect intermediate decoder behaviors, our approach differs in two key aspects. \textit{First}, instead of projecting image patches into the text token space, we directly decode the last hidden state at each decoder layer. This choice allows us to capture the model’s \textit{internal decision flow}, rather than its cross-modal alignment. Second, decoding the final hidden state per layer reflects the same mechanism used by the final decoder layer to generate the next token. Consequently, our method reveals the evolving token-level predictions (\ie, the thoughts) formed throughout the decoder layers, rather than merely tracing what the model has seen so far. This provides a more faithful characterization of how the model incrementally converges toward its final output.

\subsubsection{Feature Extraction}
\label{sec:feat_extract}
To predict hallucination with a lightweight classifier, we construct a feature vector comprising several indicators of internal uncertainty.

\textbf{Overthinking Score.}
Among these, the \textit{Overthinking Score} serves as a primary feature.
We define the Overthinking Score as the number of unique top-1 tokens emitted across $L$ decoder layers, reflecting the diversity during reasoning:
\begin{equation}
S_{\text{OT}} = 
\frac{|\{x_\ell \mid \ell \in [1, L]\}|}{L}
\cdot
\frac{\sum_{\ell=1}^{L} H_\ell}{L},
\end{equation}
where $x_\ell$ denotes the top-$1$ token, and $H_\ell$ denotes the entropy at layer $\ell$, quantifying the uncertainty of its token distribution:
\begin{equation}
H_\ell = -\sum_{v \in V} p_\ell(v) \log p_\ell(v), 
\quad \ell = 1, \dots, L.
\end{equation}
The first term measures the variability of the top-$1$ predictions across layers, while the second term captures the average uncertainty of those layers.
A high $S_{\text{OT}}$ therefore indicates that the model entertains too many competing beliefs, which characterizes ``overthinking".
We show that $S_{\text{OT}}$ is a strong indicator of hallucination in Fig.~\ref{fig:shap}, with more information in Supp.

\textbf{Attention Score.}
In addition, we include attention-based features to capture how strongly the next token attends to visual and textual evidence.
Attention distributions reflect how information flows between modalities during token generation.
Although we empirically show that attention is not a reliable indicator of hallucination for strong-prior scenes, it can still provide useful signal in general.
We include a more detailed analysis in the Supp.

\textit{Next-Token to Image Attention.}
Let $\mathcal{I}$ denote the set of image token indices and $A_{\ell}^{(h)}$ the attention matrix of head $h$ in layer $\ell$.
For each layer, we compute the average image attention received by the next token $x_t$ as:
\begin{equation}
    \alpha_{\ell}^{\text{img}}
    = \frac{1}{|\mathcal{I}|} \sum_{i \in \mathcal{I}} \max_{h} A_{\ell}^{(h)}(t,i),
\end{equation}
where $A_{\ell}^{(h)}(t,i)$ denotes the attention weight from $x_t$ to the $i$-th image token.
This metric quantifies how much the predicted token relies on visual grounding.
Lower values indicate that the model underutilizes image features, which has been shown to be indicative of hallucination~\cite{svar, metatoken}.

\textit{Next-Token to Text Attention.}
Similarly, we measure how strongly $x_t$ attends to previous textual context tokens.
Let $\mathcal{T}$ denote the set of text token indices before $t$.
The average text attention at layer $\ell$ is defined as:
\begin{equation}
    \alpha_{\ell}^{\text{text}}
    = \frac{1}{|\mathcal{T}|} \sum_{i \in \mathcal{T} \setminus \{t\}} \max_{h} A_{\ell}^{(h)}(t,i).
\end{equation}
High text attention indicates that the model is relying more on linguistic priors rather than visual evidence, which correlates with object hallucination.
By incorporating both image and text attention into the feature vector, we can capture both visual grounding and linguistic bias for each token.

\textbf{Feature Vector.}
To predict hallucination for the generated token $x_t$, we aggregate all these indicators into a unified feature representation.
Let $\mathbf{H} = (H_1, \dots, H_L)$, 
$\boldsymbol{\alpha}^{\text{img}} = (\alpha_{1}^{\text{img}}, \dots, \alpha_{L}^{\text{img}})$, and 
$\boldsymbol{\alpha}^{\text{text}} = (\alpha_{1}^{\text{text}}, \dots, \alpha_{L}^{\text{text}})$ 
denote the entropy, image attention, and text attention vectors across all $L$ layers, respectively.
Together with the Overthinking Score $S_{\text{OT}}$, these quantities encode both the uncertainty and multi-modal reasoning behavior.
The final feature vector is constructed by concatenating all components:
\begin{equation}
    \phi(x_t) = [\,S_{\text{OT}}  \,\mathbin{||}\, \mathbf{H}\,\mathbin{||}\, \boldsymbol{\alpha}^{\text{img}} \,\mathbin{||}\, \boldsymbol{\alpha}^{\text{text}}\,],
\end{equation}
where $\mathbin{||}$ denotes vector concatenation.

\subsubsection{Hallucination Detection Model}
\label{sec:detection_model}
Given the feature vector $\phi(x_t)$, the final step is to classify whether the generated token $x_t$ corresponds to a real object or a hallucinated one.
We train a lightweight binary classifier $M$  on token-level annotations, where each training sample consists of a feature vector and a ground-truth label indicating whether the object exists in the image.
The classifier maps the feature representation to a hallucination probability:
\begin{equation}
    \hat{y}_t = M(\phi(x_t)),
\end{equation}
where $\hat{y}_t$ represents hallucination prediction.


\begin{figure*}
    \centering
    \includegraphics[width=0.9\linewidth]{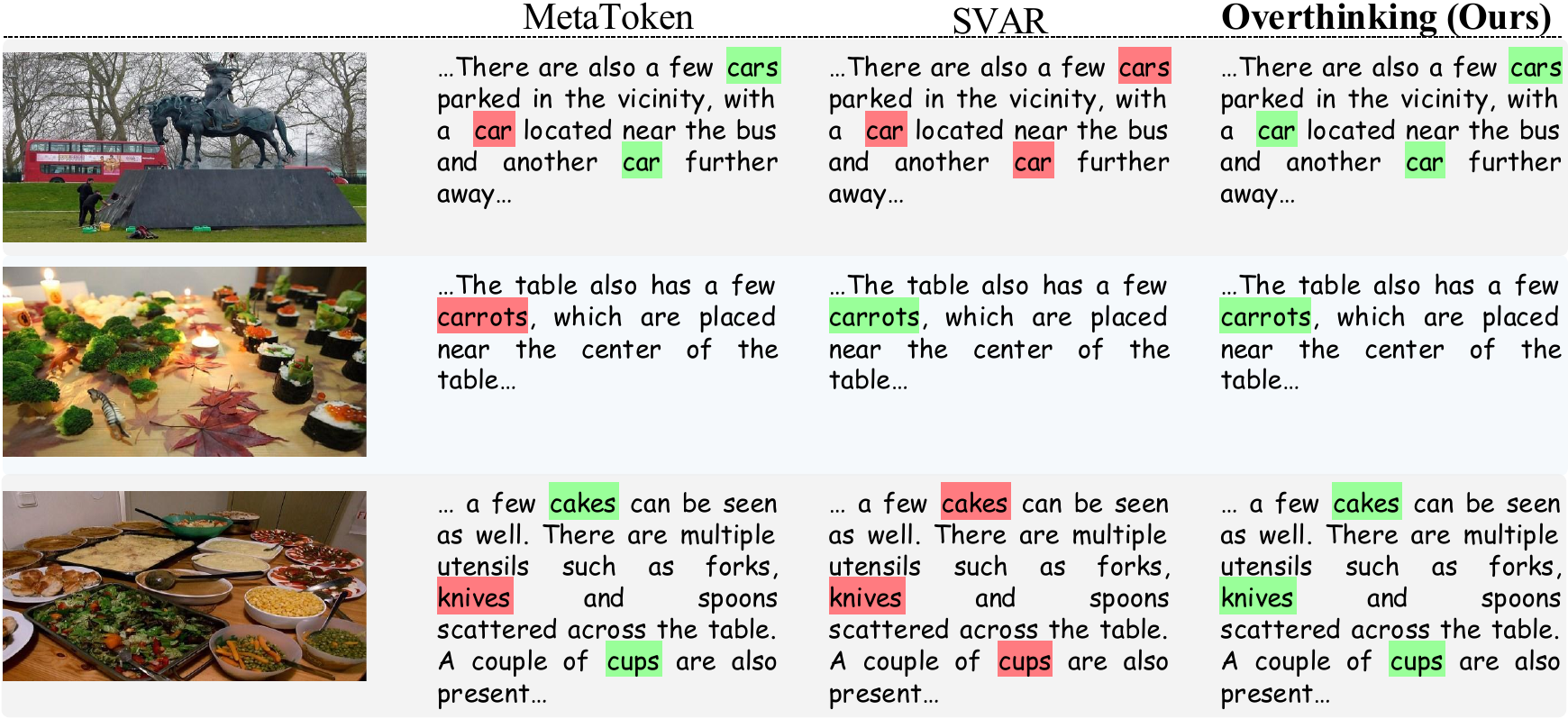}
    \caption{\textbf{Qualitative comparison of our method with MetaToken and SVAR.}
    Each row presents an example, and each column shows the prediction made by each method. Correct hallucination detections are highlighted with \colorbox{mintgreen}{green} boxes, while missed detections are shown in \colorbox{rose}{red}.}
    \label{fig:quali_results}
\end{figure*}

\section{Experimental Setup}
\label{sec:results}

\textbf{Models.} We evaluate three popular VLMs including LLaVA-1.5-7B~\cite{llava}, Qwen3-VL-4B~\cite{qwen3}, and Gemma-3-4B~\cite{gemma} to understand the generalization of our method. \\
\textbf{Dataset.} Following~\cite{opera}, we conduct experiments on 4000 random image-annotations from the MSCOCO 2014 validation dataset~\cite{mscoco}. 
The selected set is further segregated into training and test subsets, with 10\% allocated for testing and the remaining for training. Each VLM is separately prompted to describe the image (i.e., ``Describe this image.") with a maximum token limit of 1024 and labelled at the object token level using GPT-4o~\cite{achiam2023gpt} to classify the object tokens as real or hallucinated. Additionally, we evaluate the out-of-distribution (OOD) performance of our trained detectors using AMBER dataset~\cite{wang2023llm}.

\noindent\textbf{Implementation Details.}
Our method uses three lightweight classifiers, including Logistic Regression (LR), Gradient Boosting Classifier (GB), and Multi-layer Perceptron (MLP). The Logistic Regression model is trained for 2000 iterations with the L-BFGS solver. The Gradient Boosting Classifier has 200 estimators with a maximum depth of 10 and is trained with a learning rate of 0.1. Finally, the MLP classifier consists of 128 hidden layer units followed by ReLU activation and is trained for 2000 epochs with a learning rate of 0.01. Hyperparameter optimization details can be found in Supp.\\
\noindent\textbf{Tasks.} Our main task is token-level object hallucination detection via prefix prompting, following the setup in~\cite{svar}.

\noindent\textbf{Benchmarks and Metrics.}
Following~\cite{metatoken, svar, halloc}, for \textit{Hallucination Detection}, we report AUC, AP, and F1-score under dataset imbalance.

\noindent\textbf{Baselines.} 
We employ three recent hallucination detection techniques as the detection baselines, including SVAR~\cite{svar}, HalLoc~\cite{halloc}, and MetaToken~\cite{metatoken}. SVAR uses attention values from the intermediate layers for features to train an MLP-based hallucination detector. HalLoc uses the last hidden state representation of the image from the CLIP encoder as a feature for the detection module, consisting of the VisualBERT encoder followed by linear classification heads. MetaToken utilizes final layer entropy, image attention, and several other probability distribution features to train two different detectors, including Logistic Regression and Gradient Boosting classifiers. \\

\section{Results}
\begin{figure}
    \centering
    \includegraphics[width=1\linewidth]{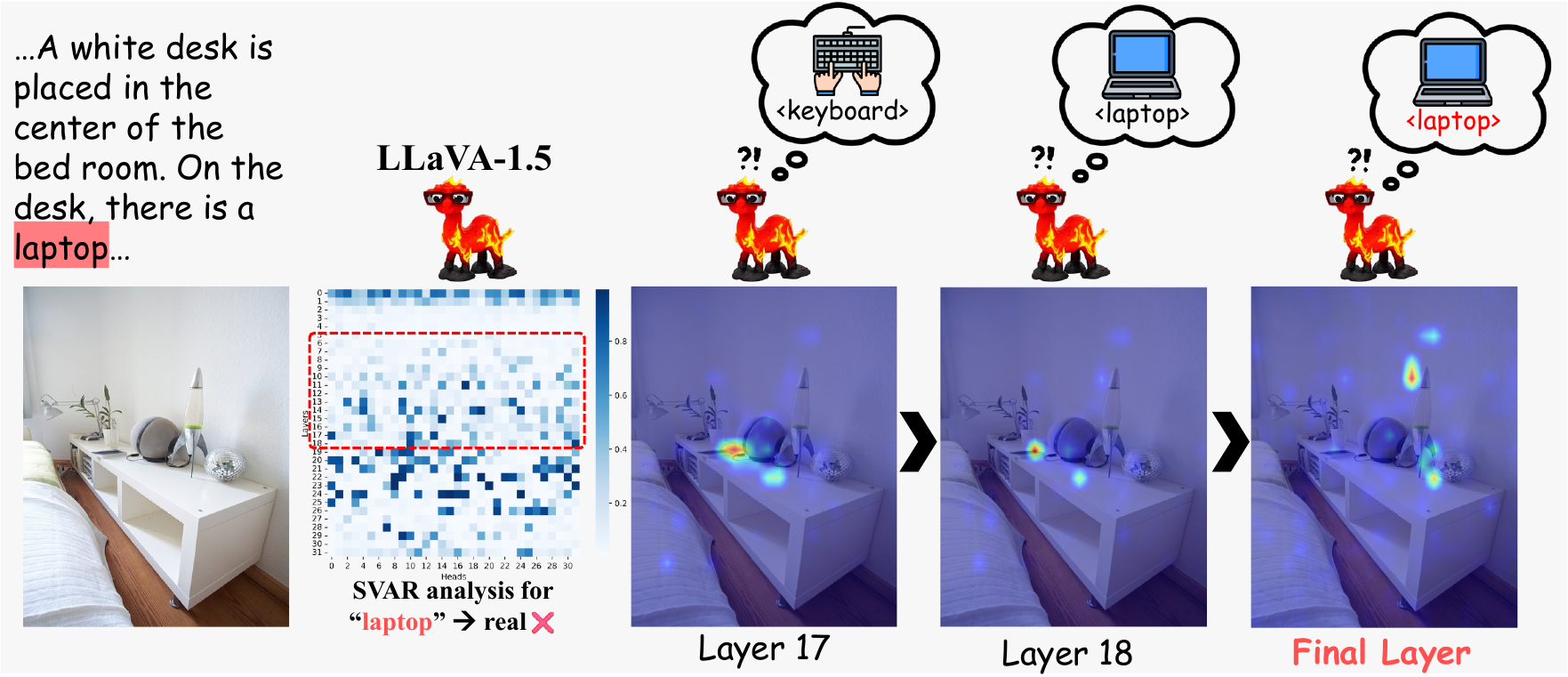}
    \caption{Illustration of an instance where an attention-based mechanism (SVAR) can highly attend to inconsistent objects and fails to detect the hallucinated token \colorbox{rose}{laptop}. LLaVA highly attends to the book object and leads to confused thought keyboard $\rightarrow$ laptop $\rightarrow$ laptop. Our method successfully captures the hallucination by tracing this thought process.}
    \label{fig:attention_fail}
\end{figure}

\noindent\textbf{Hallucination Detection}
\cref{tab:det_auc_metrics} presents the result of hallucination detection against the selected baselines across 3 VLMs under MSCOCO dataset. On average, the MLP variant of our method yields the highest AUC (87.33\%), followed by the GB variant (87.30\%). In terms of AP, the GB variant achieves the highest (61.54\%), followed by the MLP variant (58.12\%). Additionally, \cref{tab:amber_result} indicates that our method obtains the best generalization in the out-of-distribution (OOD) AMBER dataset with the GB variant having the superior performance.
\cref{fig:attention_fail} demonstrates the case where existing methods such as 
SVAR, 
fails in detection by attending to spurious region and assigning high attention magnitude, when the VLM has high attention on the book object resembling a laptop. Our method overcomes this limitation by leveraging the internal thought process keyboard $\rightarrow$ laptop $\rightarrow$ laptop.
Qualitative results are illustrated in Fig.~\ref{fig:quali_results}. (More in Supp.)

\begin{table*}[htbp]
\centering
\small
\setlength{\tabcolsep}{6pt}
\begin{threeparttable}
\caption{Object hallucination detection performance (in \%) on MSCOCO dataset  measured using AUC, AP and F1 across three different VLMs.}
\label{tab:det_auc_metrics}
\begin{tabular}{l *{4}{ccc}}
\toprule
& \multicolumn{3}{c}{\textbf{LLaVA-1.5}} & \multicolumn{3}{c}{\textbf{Gemma-3}} & \multicolumn{3}{c}{\textbf{Qwen3-VL}} & \multicolumn{3}{c}{\textbf{Avg.}}\\
\cmidrule(lr){2-4}\cmidrule(lr){5-7}\cmidrule(lr){8-10}\cmidrule(lr){11-13}
\textbf{Method} & AUC & AP & F1 & AUC & AP & F1 & AUC & AP & F1  & AUC & AP & F1\\
\midrule
SVAR~\cite{svar}            & 85.12 & 50.68 & 69.35 & 74.11 & 32.75 & 47.84 & 75.56 & 35.69 & 50.20 & 78.26 & 39.71  & 55.80\\
HalLoc~\cite{halloc}          & 80.38 & 53.61 & 73.68 & 79.27 & 49.96 & 67.11 & 83.85  & \underline{59.50} & \textbf{74.75} & 81.17 & 54.36 & 71.85 \\
MetaToken (LR)~\cite{metatoken}   & 85.41 & 52.59 & 72.88 & 73.91 & 28.78 & 48.27 & 79.49 &  27.60 & 47.27 & 79.60 & 36.32 & 56.14\\
MetaToken (GB)~\cite{metatoken}   & 88.95 & 61.03 & \underline{75.95} & 77.23 & 34.75 & 67.15 & 84.21 &  38.06 & \underline{74.43} & 83.46 & 44.61 & 72.51 \\
MetaToken (MLP)~\cite{metatoken}   & 86.81 & 55.72 & 73.89 & 78.40 & 33.63 & 60.03 & 86.29 &  47.00 & 68.03 & 83.83 & 45.45 & 67.32 \\
\midrule
\textbf{Ours (LR)}        & 86.85 & 59.16 & 72.88 & 76.68 & 31.20 & 67.53 & 77.85 & 37.31  & 55.90 & 80.46 & 42.56 & 65.44 \\
\textbf{Ours (GB)}        & \underline{89.66} & \underline{63.04} & \textbf{78.95} & \textbf{85.59}  & \textbf{59.89}  & \textbf{74.54} & \underline{86.65} & \textbf{61.69}  & \underline{74.43} & \underline{87.30} & \textbf{61.54} & \textbf{75.97} \\
\textbf{Ours (MLP)}       & \textbf{89.73} & \textbf{63.81} & 75.37 & \underline{85.38} & \underline{56.84} & \underline{72.07} & \textbf{86.89} & 53.72 & 71.15 & \textbf{87.33}  & \underline{58.12} & \underline{72.86} \\
\bottomrule
\end{tabular}
\end{threeparttable}
\end{table*}

\begin{table}[]
    \centering
    \begin{tabular}{lccc}
    \toprule
        \textbf{Method} &  AUC  & AP & F1 \\
        \midrule
         SVAR~\cite{svar} & 79.51 & 12.78 & 56.87  \\
         HalLoc~\cite{halloc} & 50.00 & 2.27 & 49.44 \\
         MetaToken (LR)~\cite{metatoken} & 79.50 & 11.55 & 54.76 \\
         MetaToken (GB)~\cite{metatoken} & 82.15 & 23.48 & \underline{65.54} \\
         MetaToken (MLP)~\cite{metatoken} & 79.41 & 14.04 & 58.54 \\
         \midrule
         \textbf{Ours (LR)} & 70.76 & 7.97 & 53.92 \\
         \textbf{Ours (GB)} & \textbf{86.11} & \textbf{36.65} & \textbf{71.58} \\
         \textbf{Ours (MLP)} & \underline{83.28} & \underline{30.91} & 59.76 \\
         \bottomrule
    \end{tabular}
    \caption{OOD object hallucination detection performance (in \%) on AMBER dataset and LLaVA-1.5 VLM measured using AUC, AP and F1.}
    \label{tab:amber_result}
\end{table}
\section{Ablation Study}
\begin{figure}
    \centering
    \includegraphics[width=0.9\linewidth]{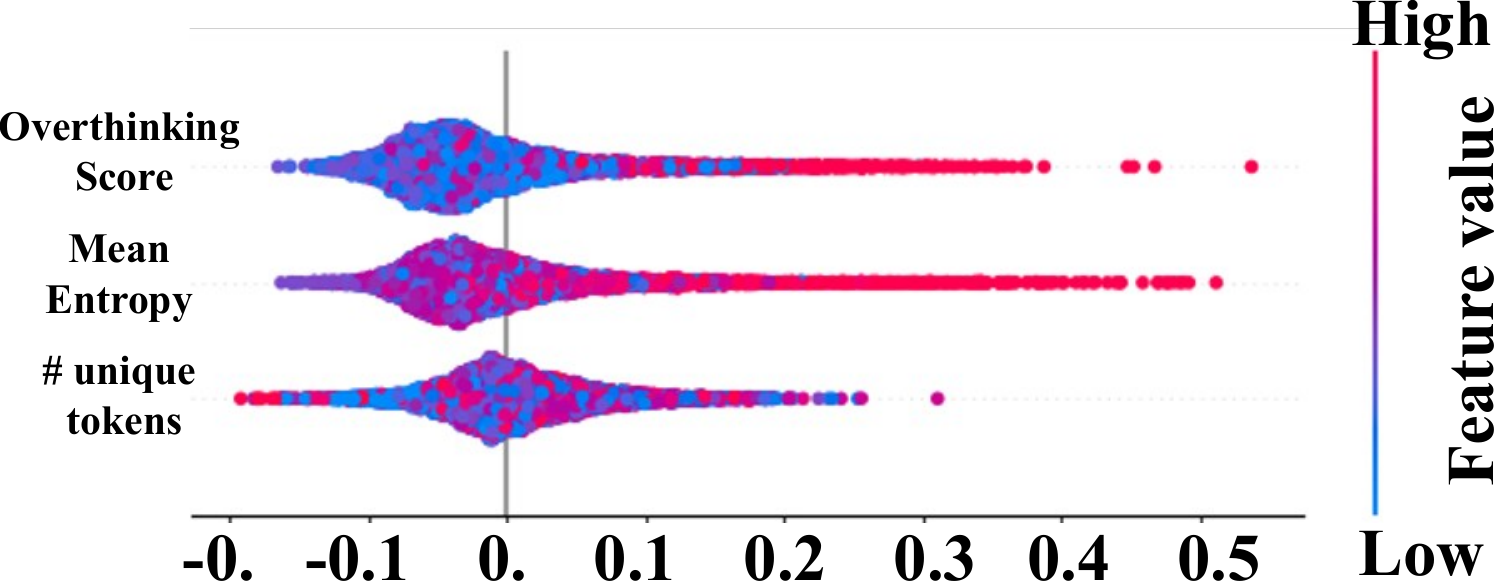}
    \vspace{-0.5em}
    \caption{SHAP analysis indicating that both mean entropy and number of unique token are informative for detecting hallucination, while the compound Overthinking Score offers the strongest overall contribution.}
    \label{fig:shap}
\end{figure}

\begin{figure}
    \centering
    \includegraphics[width=0.9\linewidth]{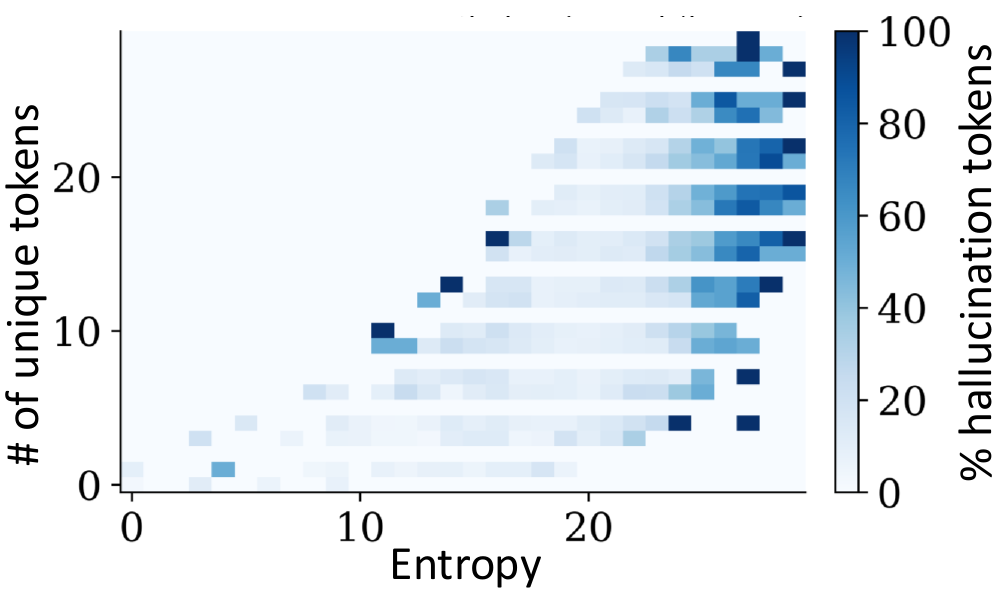}
    \vspace{-0.5em}
    \caption{Hallucinations concentrate in regions of both
high entropy and token diversity (i.e., with MSCOCO \& LLaVA-1.5).}
    \label{fig:mult}
\end{figure}

\textbf{Feature Importance.}
We ablate different indicator in the feature vector. Among the indicators, Overthinking Score brings the most significant improvement in AUC ($86.58\% \rightarrow 89.73\%$). We include more details in Supp.\\
\textbf{Overthinking Components.}
To understand the contribution of individual components in the overthinking score, we estimate their SHAP (SHapley Additive exPlanations)~\cite{shap}. As shown in \cref{fig:shap}, both Mean Entropy and Unique Token Count positively contribute to hallucination prediction, showing that uncertainty and hypothesis diversity are informative cues. Yet, the Overthinking Score provides a clearer, more stable signal, confirming the benefit of combining both effects in a single metric. Furthermore, the higher density in the top right corner of the scatter plot shown in \cref{fig:mult} indicates that the hallucination rate (in \%) is proportional to the product of these two components.
\\
\textbf{Layer-wise importance.}
We analyze the importance of including features from all layers for hallucination detection by training an MLP classifier using layer subsets. Since the SVAR method discusses the importance of middle layers (5-18) and MetaToken utilize the final layer entropy, we group them as [0-4], [5-18], [19-31], and finally use only the last layer. As shown in \cref{tab:layers_selection},
later layers set [19-31] shows better performance than earlier layers.
Utilizing \textit{all layers} gives the best F1 and AUC scores, confirming that all layers are beneficial for hallucination detection.
\\




\begin{table}[]
    \centering
    \begin{tabular}{cccccc}
    \toprule
        \textbf{Layers} & \textbf{0-4} & \textbf{5-18} & \textbf{19-31} & \textbf{31} & \textbf{0-31} \\
        \midrule
        AUC & 85.14 & 87.37 & \underline{88.93} & 83.79 & \textbf{89.73} \\
        F1 & 67.67 & 71.61 & \underline{74.75} & 68.76 & \textbf{75.37} \\
    \bottomrule
    \end{tabular}
    \vspace{-0.5em}
    \caption{AUC and F1 scores using different layer subsets for hallucination detection.}
    \label{tab:layers_selection}
\end{table}

\section{Conclusion}
We introduce a novel perspective on the cause of hallucination in Vision Language Models. We show that hallucination originates from early-layer activations that propagate confounding factors through depth, a process we term ``confounder propagation". This internal drift is not visible to final-layer uncertainty methods. As object hypotheses diversify by layers, confounders accumulate and trigger overthinking that leads to hallucination. Capturing this  with Overthinking Score, we effectively detect hallucinations across multiple VLMs. Future works include extension to broader reasoning failures and developing mitigation strategies guided by confounders dynamics.
{
    \small
    \bibliographystyle{ieeenat_fullname}
    \bibliography{main}
}
\clearpage
\setcounter{page}{1}
\maketitlesupplementary


\section{Hyperparameters Optimization}

The hyperparameters of GB and MLP variants of the detector are finetuned using Grid Search on a selected set of parameters. The training dataset is split into a train and validation set with a 10\% allocation for the validation set. The models are finetuned to improve the F1-score on the validation set.

\textbf{GB Optimization}: The optimal value of the number of estimators, learning rate (LR) and maximum depth parameters are chosen from the set \{100, 200, 300\}, \{0.1, 0.05, 0.01\}, and \{3, 5, 10\} respectively. The performance of the model under different hyperparameter combinations is shown in \cref{tab:hyperparams_gb}.

\begin{table}[]
    \centering
    \begin{tabular}{c|c|c|c}
    \toprule
    \textbf{LR} &  \textbf{Maximum depth} & \textbf{No. of estimators} & \textbf{F1 (\%)}  \\
    \midrule
    0.1 & 3 & 100 & 71.39 \\
    0.1	& 3 & 200 & 72.43 \\
    0.1	& 3	& 300 &	72.72 \\
    0.1	& 5	& 100 &	73.59 \\
    0.1	& 5	& 200 & 74.25 \\
    0.1	& 5	& 300 & 73.06 \\
    0.1	& 10 & 100 & 73.42 \\
    \textbf{0.1}	& \textbf{10} & \textbf{200} & \textbf{74.38} \\
    0.1	& 10	& 300	& 74.19 \\
    0.05	& 3	& 100	& 71.02 \\
    0.05	& 3	& 200	& 71.59 \\
    0.05	& 3	& 300	& 72.16 \\
    0.05 & 5  & 100 & 72.40 \\
    0.05 & 5  & 200 & 72.83 \\
    0.05 & 5  & 300 & 72.81 \\
    0.05 & 10 & 100 & 73.43 \\
    0.05 & 10 & 200 & 72.87 \\
    0.05 & 10 & 300 & 74.19 \\
    0.01 & 3  & 100 & 54.53 \\
    0.01 & 3  & 200 & 66.30 \\
    0.01 & 3  & 300 & 68.97 \\
    0.01 & 5  & 100 & 58.18 \\
    0.01 & 5  & 200 & 68.28 \\
    0.01 & 5  & 300 & 71.48 \\
    0.01 & 10 & 100 & 61.18 \\
    0.01 & 10 & 200 & 68.06 \\
    0.01 & 10 & 300 & 70.18 \\
    \bottomrule
    \end{tabular}
    \caption{Results of hyperparameter tuning with GB model.}
    \label{tab:hyperparams_gb}
\end{table}

\textbf{MLP Optimization}: The optimal value of hidden units, learning rate (LR) and optimizers are chosen from the set \{32, 64, 128\}, \{0.1, 0.01, 0.001\}, and \{Adam, SGD\} respectively. The performance of the model on validation set under different hyperparameter combinations is shown in \cref{tab:hyperparams_mlp}.

\begin{table}[]
    \centering
    \begin{tabular}{c|c|c|c}
    \toprule
    \textbf{Hidden units} &  \textbf{LR} & \textbf{Optimizer} & \textbf{F1 (\%)}  \\
    \midrule
    32  & 0.1   & Adam & 69.50 \\
    32  & 0.1   & SGD  & 77.81 \\
    32  & 0.01  & Adam & 73.97 \\
    32  & 0.01  & SGD  & 73.48 \\
    32  & 0.001 & Adam & 76.04 \\
    32  & 0.001 & SGD  & 72.99 \\
    64  & 0.1   & Adam & 75.02 \\
    64  & 0.1   & SGD  & 77.02 \\
    64  & 0.01  & Adam & 73.94 \\
    64  & 0.01  & SGD  & 78.09 \\
    64  & 0.001 & Adam & 75.37 \\
    64  & 0.001 & SGD  & 72.82 \\
    128 & 0.1   & Adam & 72.77 \\
    128 & 0.1   & SGD  & 74.66 \\
    128 & 0.01  & Adam & 75.66 \\
    \textbf{128} & \textbf{0.01}  & \textbf{SGD}  & \textbf{78.31} \\
    128 & 0.001 & Adam & 74.91 \\
    128 & 0.001 & SGD  & 73.65 \\
    \bottomrule
    \end{tabular}
    \caption{Results of hyperparameter tuning with MLP model.}
    \label{tab:hyperparams_mlp}
\end{table}





\section{Semantic Alignment Calculation}


\paragraph{Maximum Semantic Alignment} We employ the sentence-transformer all-MiniLM-L6-v2~\cite{sentence_transformer} to embed tokens decoded from different VLM layers. For each generated token, we compute the cosine similarity between its final-layer embedding and the embeddings of tokens from all intermediate layers. The semantic alignment is defined as the maximum similarity across layers, as shown in \cref{alg:semantic_sim}.

\begin{algorithm}
\caption{Maximum Semantic Alignment Computation}
\label{alg:semantic_sim}
\KwIn{Top-1 tokens at each layer $x_1, \ldots, x_L$ and sentence transformer $T$}
\KwOut{Semantic alignment value $S_{\text{align}}$}

\textbf{Initialize:} $S_{\text{align}} \gets 0$\;

Embed the final-layer token: $x_L^{\text{emb}} \gets T(x_L)$\;

\ForEach{$x_i \in \{x_1, \ldots, x_{L-1}\}$}{
    \If{$x_i \neq x_L$}{
        Embed token: $x_i^{\text{emb}} \gets T(x_i)$\;
        Compute cosine similarity: 
        $S \gets \text{cosine\_similarity}(x_L^{\text{emb}}, x_i^{\text{emb}})$\;
        \If{$S > S_{\text{align}}$}{
            $S_{\text{align}} \gets S$\;
        }
    }
}
\Return $S_{\text{align}}$\;
\end{algorithm}

\begin{figure}
    \centering
    \includegraphics[width=1\linewidth]{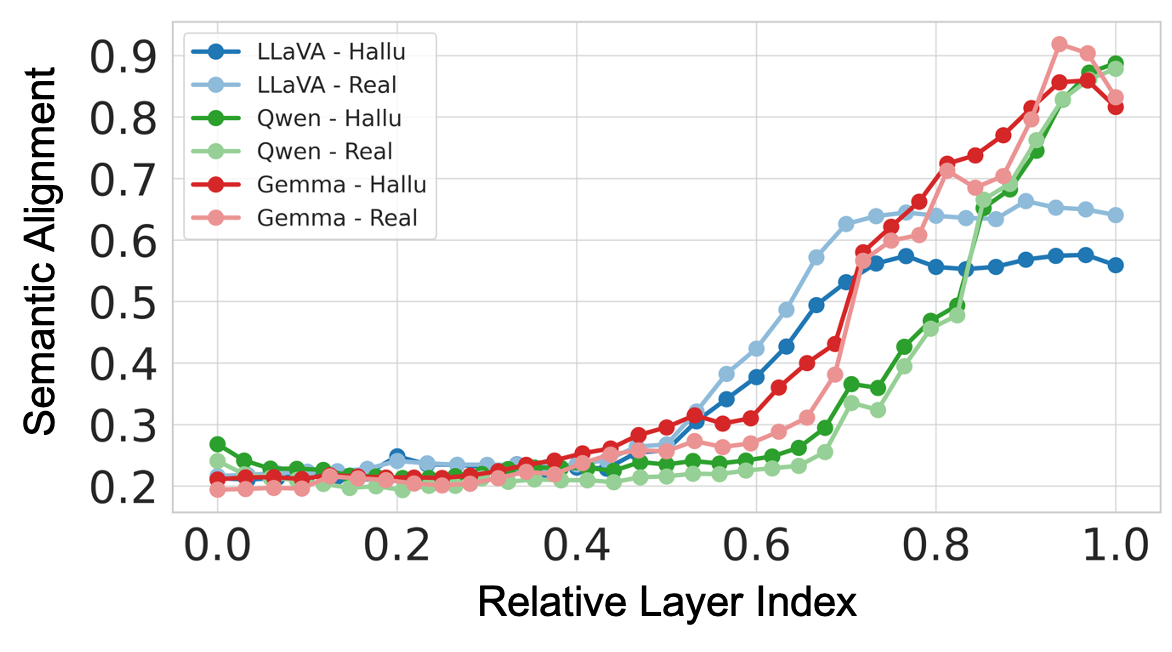}
    \caption{Semantic alignment of hallucinated (hallu) and real (real) token instances across the layers (relative index) for LLaVA-1.5, Gemma-3 and Qwen3-VL models.}
    \label{fig:semantic_align_layer}
\end{figure}

\paragraph{Layer-wise Semantic Alignment}
We analyze how intermediate layers influence the final prediction by computing the cosine similarity between the final-layer token (both hallucinated and real) and the tokens decoded at each layer. The layer-wise alignment curves for LLaVA-1.5, Gemma-3, and Qwen3-VL in \cref{fig:semantic_align_layer} show that semantic similarity becomes high from the middle layers onward, indicating that these intermediate representations strongly shape  final outputs.

\begin{table*}[htbp]
\centering
\small
\setlength{\tabcolsep}{4pt}
\begin{threeparttable}
\caption{Impact of overthinking score integration in existing hallucination detection methods.}
\label{tab:ot_addition}
\begin{tabular}{l *{2}{cc}}
\toprule
& \multicolumn{2}{c}{\textbf{w/o. Overthinking Score}} & \multicolumn{2}{c}{\textbf{w. Overthinking Score}} \\
\cmidrule(lr){2-3}\cmidrule(lr){4-5}
\textbf{Method} & AUC  & F1 & AUC & F1 \\
\midrule
SVAR~\cite{svar} & 85.12  & 69.35 & \textbf{86.67}  & \textbf{75.06} \\
HalLoc~\cite{halloc}         & 80.38  & 73.68 & \textbf{88.53}  & 72.97 \\
MetaToken (LR)~\cite{metatoken}   & 85.41  & 72.88 & \textbf{87.67}  & 70.09\\
MetaToken (GB)~\cite{metatoken}   & 88.95  & 75.95 & \textbf{89.15}  & \textbf{76.14} \\
MetaToken (MLP)~\cite{metatoken}   & 86.81  & 73.89 & \textbf{89.23}  & 72.35 \\
\bottomrule
\end{tabular}
\end{threeparttable}
\end{table*}

\begin{figure}
    \centering
    \includegraphics[width=1\linewidth]{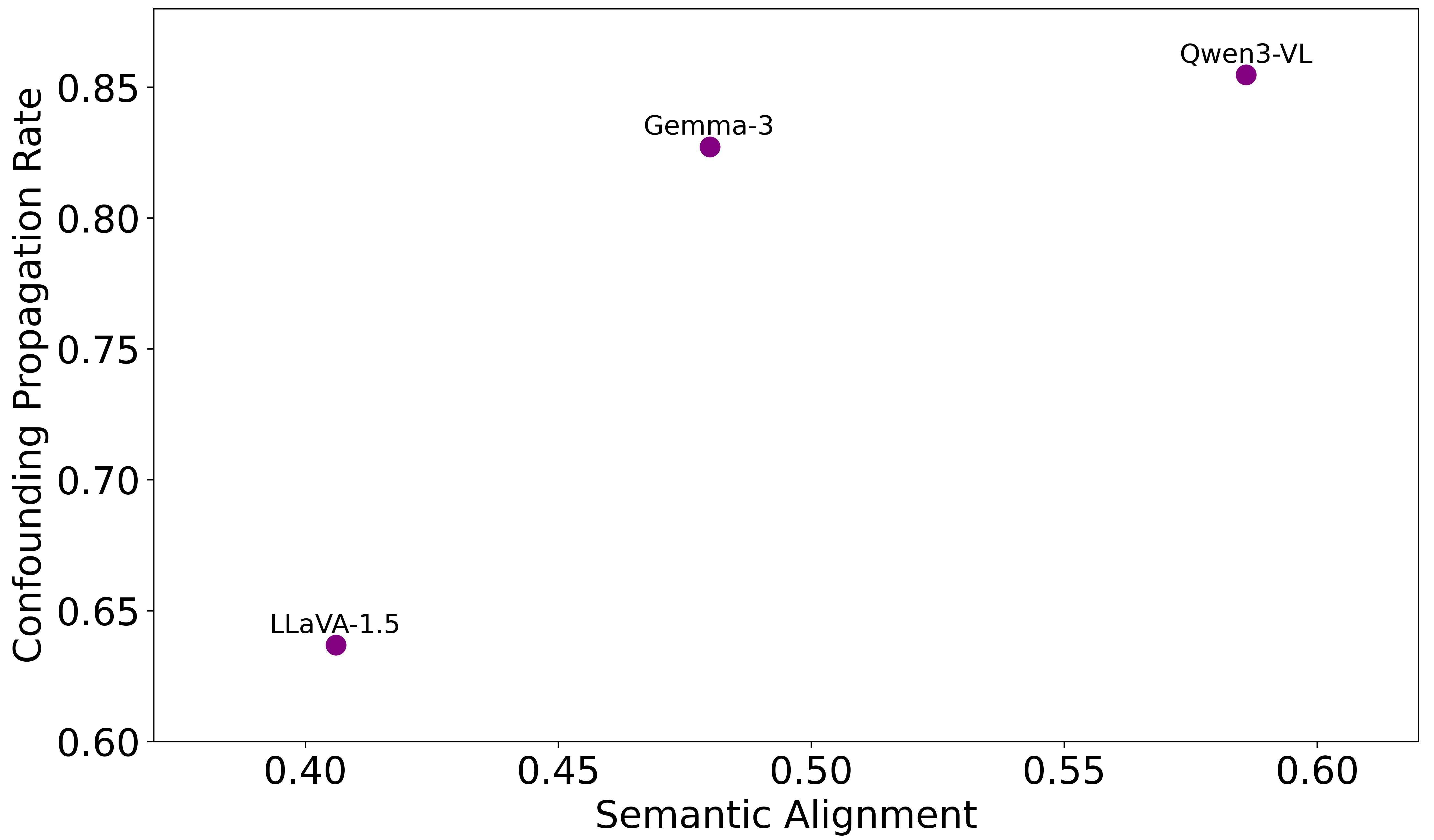}
    \caption{Semantic Alignment vs Confounding Proppagation Rate in LLaVA-1.5, Qwen3-VL and Gemma-3 models.}
    \label{fig:align_vs_conf}
\end{figure}
\section{Strong Scene-prior Examples Selection}

We use the \texttt{facebook/bart-large-mnli} model~\cite{bart} solely as an embedding encoder. The generated image description is embedded and compared (via cosine similarity) against the embeddings of all predefined scene labels (SCENE\_LABELS = [
    ``beach", ``playground", ``forest", ``park", ``street", ``mountains", ``desert",
    ``kitchen", ``living room", ``bedroom", ``office", ``classroom", ``outdoors", ``indoors", 
     ``city", ``festival", ``space", ``dining room", ``ocean", ``lake", 
     ``river"
].). The label with the highest similarity is treated as the inferred scene context, which allows us to estimate scene priors without requiring ground-truth scene annotations.
A sample is marked as having a strong scene prior when the generated object token shows similarity above 0.6 with this inferred (maximum-similarity) scene label.
The process is detailed in \cref{alg:strong_scene_prior}.

\begin{algorithm}
\caption{Extract Strong Scene Prior Cases}
\label{alg:strong_scene_prior}
\KwIn{Dataset $D$, scene extraction model $S$, sentence transformer $T$}
\KwOut{Filtered dataset $D'$}

\textbf{Initialize:} $D' \gets \{\}$\;
\textbf{Set} similarity threshold $t \gets 0.6$\;

\ForEach{$d \in D$}{
    Extract scene label: $d_{\text{scene}} \gets S(d_{\text{description}}, \text{SCENE\_LABELS})$\;
    
    Let the next token of instance $d$ be $d_{x_t}$\;

    Embed scene and token:\\
    \hspace{1em} $d_{\text{scene}}^{\text{emb}} \gets T(d_{\text{scene}})$\;
    \hspace{1em} $d_{x_t}^{\text{emb}} \gets T(d_{x_t})$\;

    Compute cosine similarity: 
    $c \gets \text{cosine\_similarity}(d_{\text{scene}}^{\text{emb}}, d_{x_t}^{\text{emb}})$\;

    \If{$c > t$}{
        Append $d$ to $D'$\;
    }
}
\Return $D'$\;

\end{algorithm}

\section{Attention-based Methods Failures}

\begin{figure*}
    \centering
    \includegraphics[width=1\linewidth, height=21cm]{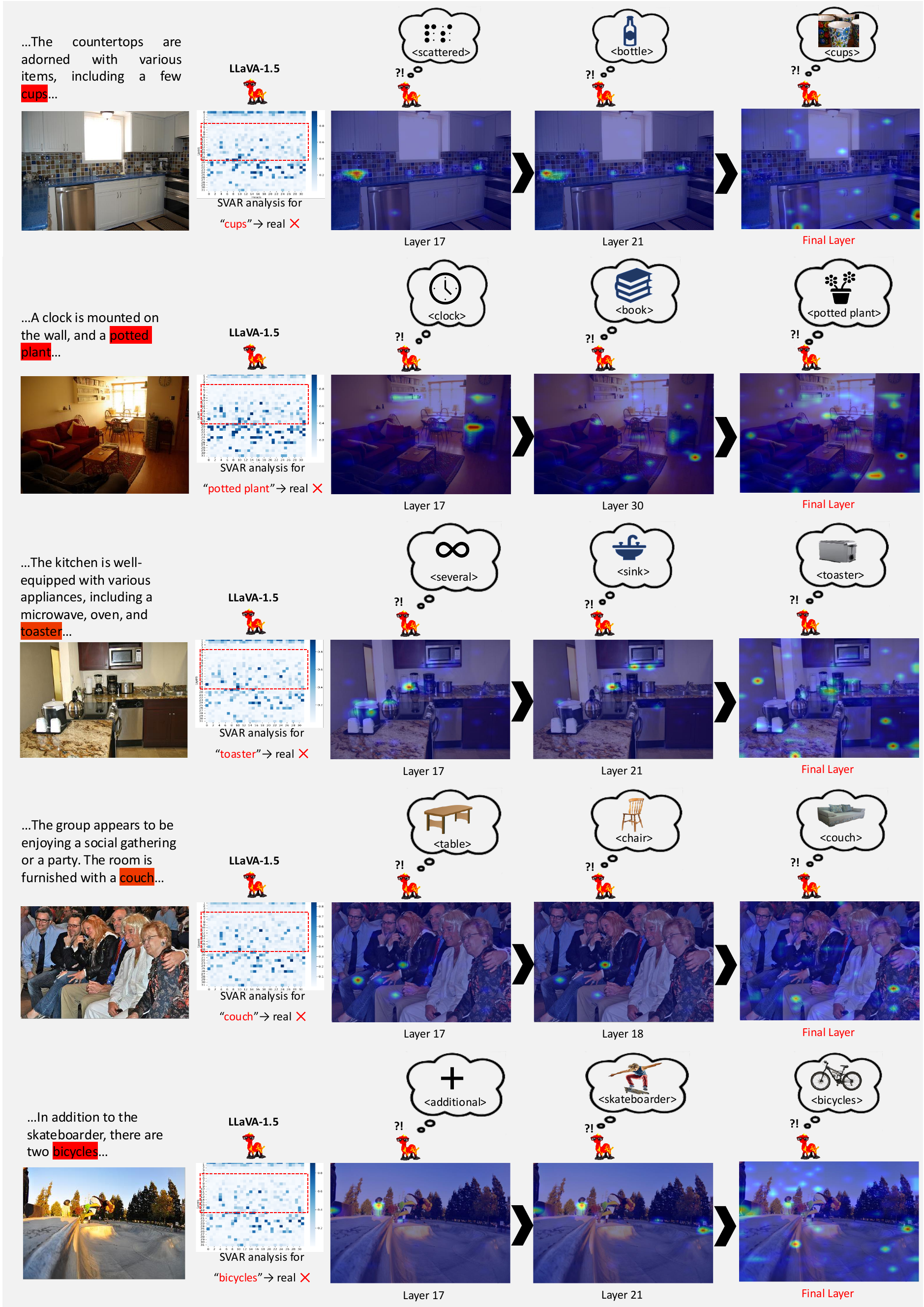}
    \caption{The above examples illustrates that SVAR fails to detect hallucinated objects when the attention values are active in the intermediate layers.}
    \label{fig:svar_examples}
\end{figure*}

We observe that attention-based methods such as SVAR fail to detect hallucinated objects when strong scene priors are present. As illustrated in \cref{fig:svar_examples}, several hallucinated tokens receive high intermediate-layer attention (layers 5–18), yet SVAR incorrectly classifies them as real because attention magnitude remains high.
To quantify this effect, we isolate strong scene-prior cases using \cref{alg:strong_scene_prior} and evaluate both SVAR and our method. As shown in \cref{tab:svar_scene_prior}, our approach achieves substantially higher AUC (86.36\%) and F1 (82.59\%), demonstrating its robustness in context-biased scenarios where SVAR fails.

\begin{table}[]
    \centering
    \begin{tabular}{cccc}
    \toprule
        \textbf{Method} & \textbf{AUC} & \textbf{F1}  \\
        \midrule
        SVAR & 76.92  & 48.86\\
        \textbf{Ours (GB)} & \textbf{86.36}  & \textbf{82.59}  \\
    \bottomrule
    \end{tabular}
    \vspace{-0.5em}
    \caption{Performance comparison (\%) of SVAR and overthinking-based approach on strong scene prior cases.}
    \label{tab:svar_scene_prior}
\end{table}

\section{S-OT Improves Other Baselines}

To highlight the benefit of the Overthinking Score, we add S-OT to the feature sets of existing detectors and compare performance. As shown in \cref{tab:ot_addition}, every method improves substantially once S-OT is included, demonstrating that S-OT provides a meaningful performance boost.
Remarkably, S-OT is only a single scalar, yet it delivers greater benefit than many multidimensional attention and entropy features, showing that it captures a uniquely informative signal missing in prior methods.


\section{Features Importance}

\paragraph{Feature Ablation} To understand the importance of each feature in our detection model, we ablate one feature at a time and observe the feature causing the largest performance drop with the GB-variant for hallucination detection in the LLaVA-1.5 baseline. Since the overthinking score is a singleton feature compared to the other features, for a fair comparison, we consider the average of image attention, text attention and entropy features across the layers. As shown in \cref{tab:ablate_drop}, the largest performance drop (83.33\% $\rightarrow$ 79.93\%) is seen when the overthinking score is removed.
\paragraph{SHAP Analysis} We estimate the significance overthinking score over other features such as image attention, text attention and entropy using the average SHAP value of the compound features on MS-COCO. The results of SHAP values for the Entropy/Text Attention/Image Attention and \textbf{Overthinking Scores} are:
0.002/0.004/0.004/\textbf{0.007} respectively.
Overthinking score has the highest importance (0.007 SHAP value) compared to other features.
\begin{table}[]
    \centering
    \begin{tabular}{lc}
        \toprule
        & \textbf{AUC} \\
        \midrule
        Full features & 83.33 \\
        \midrule
        w/o. entropy & 82.91 \\
        w/o. image attention & 82.42\\
        w/o. text attention & 80.49 \\
        w/o. Overthinking Score & \textbf{79.93} \\
    \end{tabular}
    \caption{Comparison of performance (\%) drop by ablating individual features.}
    \label{tab:ablate_drop}
\end{table}
        


\begin{table}[!t]
    \centering
    \begin{tabular}{lccc}
    \toprule
        \textbf{Method} &  Inference time (s) \\
        \midrule
        Greedy Search & 4.21 \\
        \midrule
         SVAR~\cite{svar} & 8.07  \\
         HalLoc~\cite{halloc} & 5.03 \\
         MetaToken (LR)~\cite{metatoken} & 5.35 \\
         MetaToken (GB)~\cite{metatoken} & 5.42 \\
         MetaToken (MLP)~\cite{metatoken} & 5.53 \\
         \midrule
         \textbf{Ours (LR)} & 5.61 \\
         \textbf{Ours (GB)} & 5.77 \\
         \textbf{Ours (MLP)} & 5.74 \\
         \bottomrule
    \end{tabular}
    \caption{Inference time of various hallucination detection methods.}
    \label{tab:comp_cost}
\end{table}

\section{Computational Cost}
\cref{tab:comp_cost} shows that our method incurs only 36\% additional computational cost in terms of inference time compared to the default greedy search method.

\section{Confounder Propagation Examples}
\begin{figure*}[!h]
    \centering
    \includegraphics[width=1\linewidth]{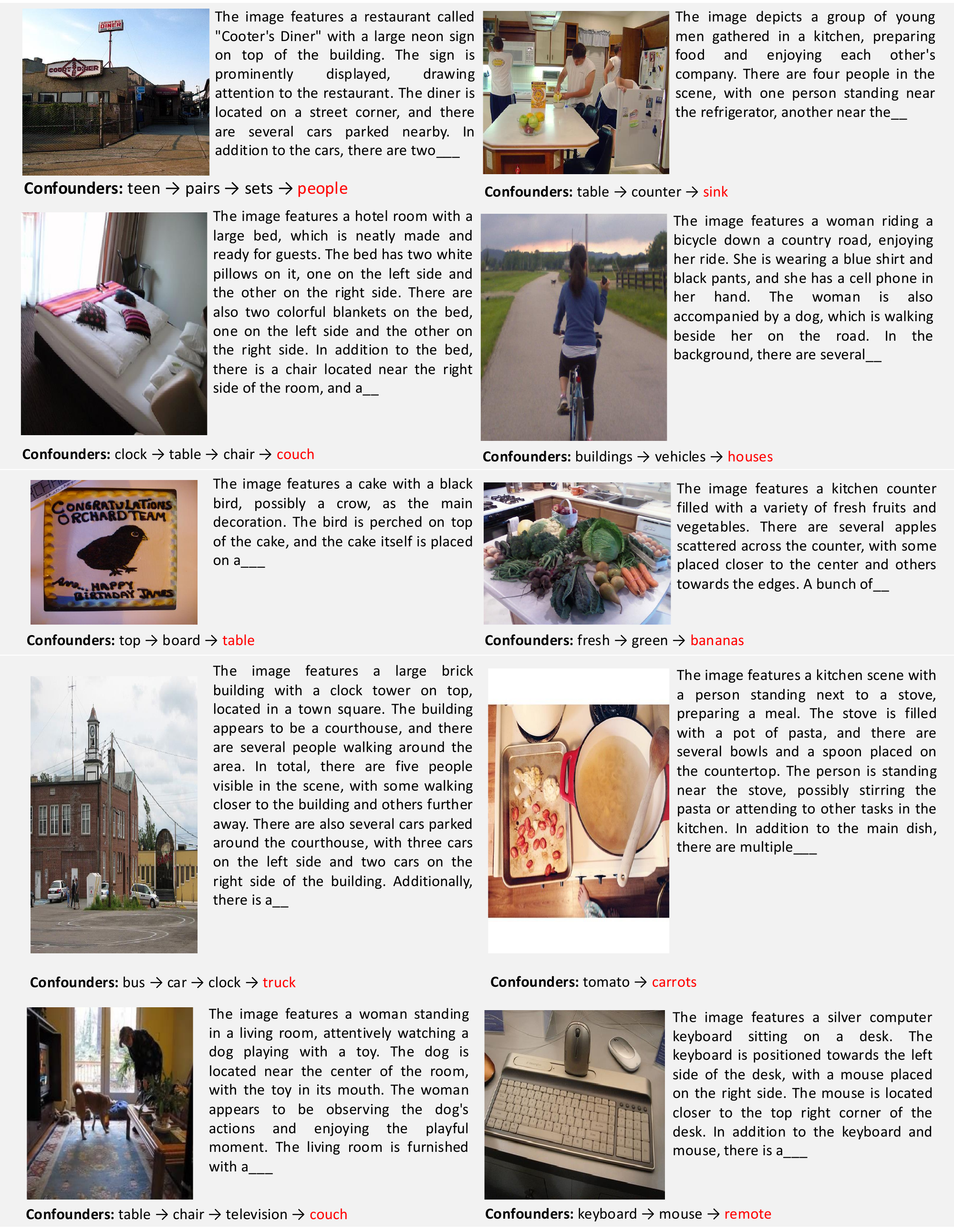}
    \caption{Some examples of confounding propagation in LLaVA-1.5.}
    \label{fig:conf_examples10}
\end{figure*}

The \cref{fig:conf_examples10} shows some examples of confounding propagation in LLaVA-1.5.

\section{Qualitative Results}
\begin{figure*}[!htbp]
    \centering
    \begin{minipage}{\textwidth}
        \includegraphics[width=\textwidth]{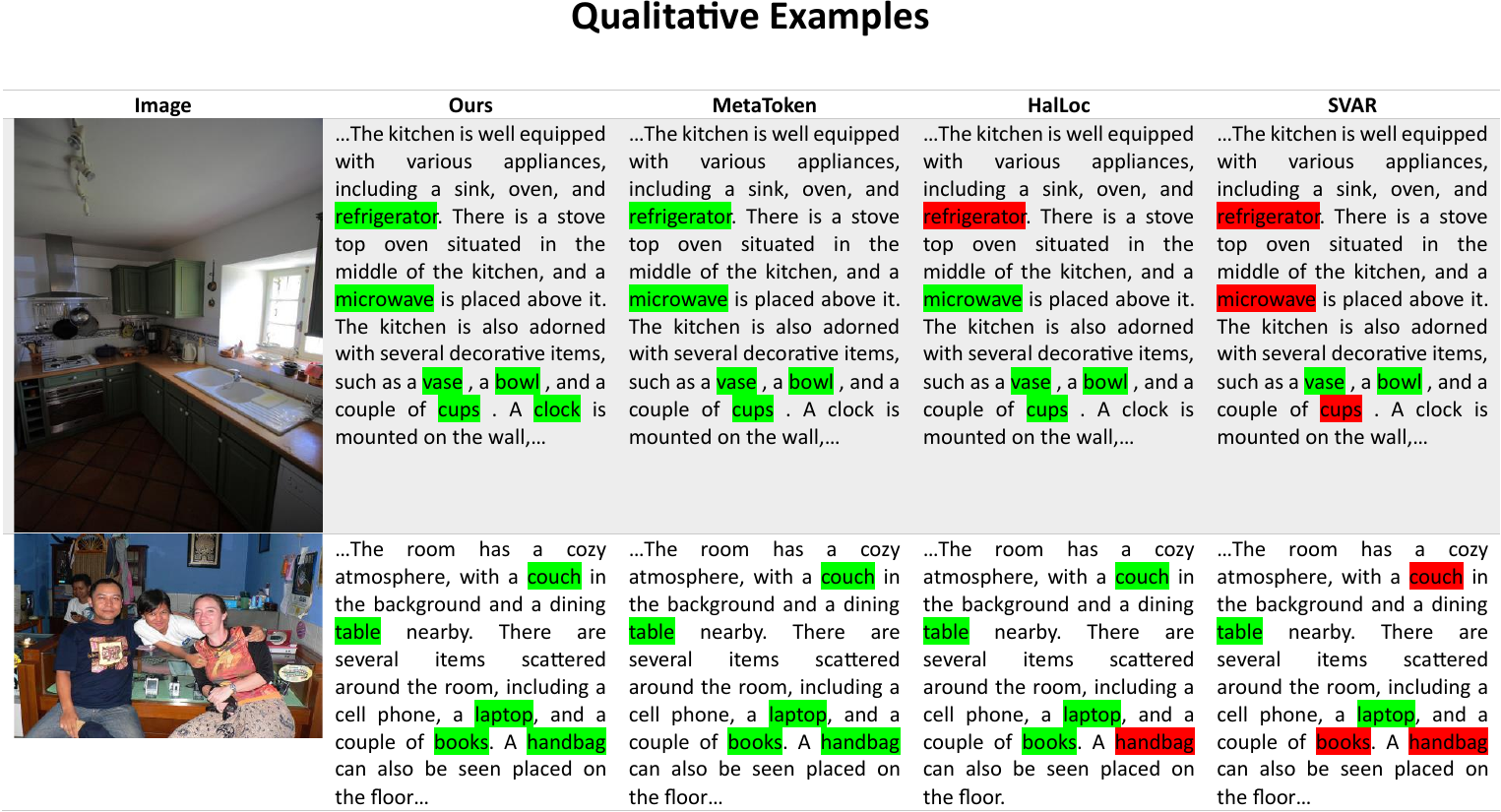}
    \end{minipage}
    \hfill
    \begin{minipage}{\textwidth}
        \includegraphics[width=\textwidth]{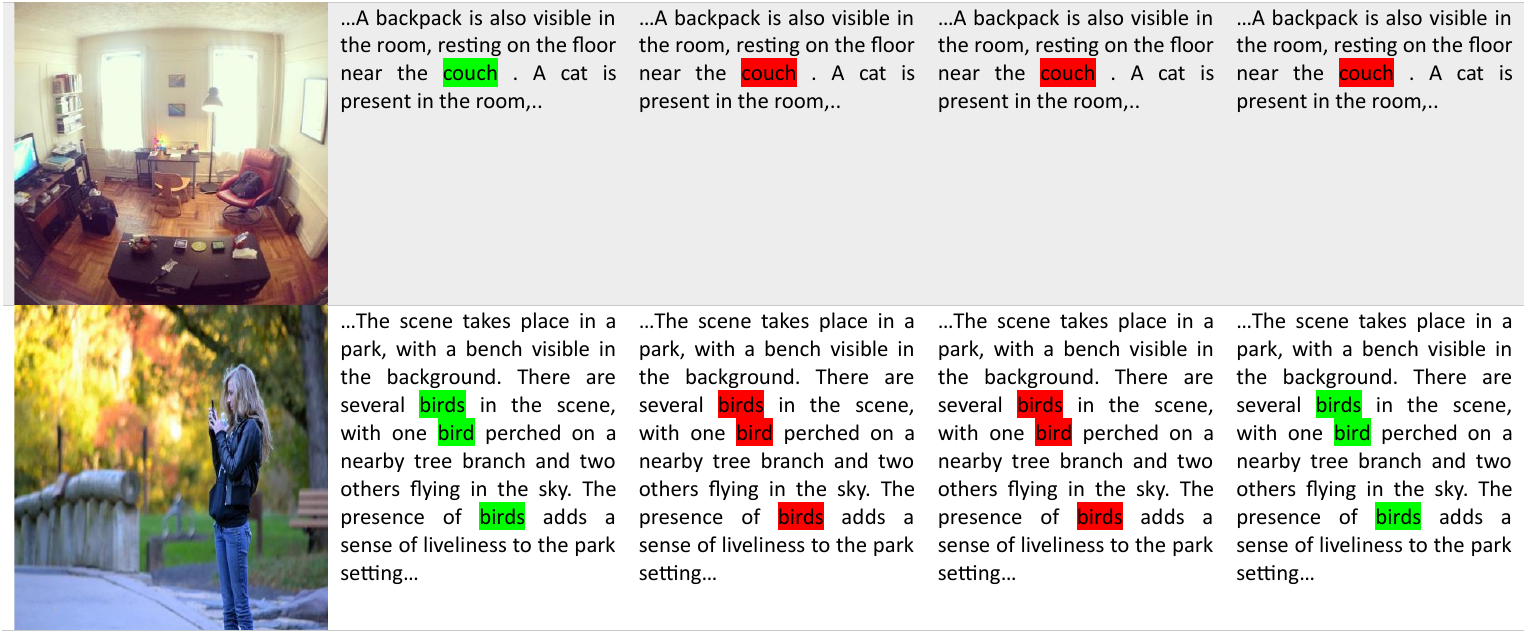}
    \end{minipage}
    \caption{Qualitative Results.}
    \label{fig:q1}
\end{figure*}

\begin{figure*}[!htbp]
    \centering
    \begin{minipage}{\textwidth}
        \includegraphics[width=\textwidth]{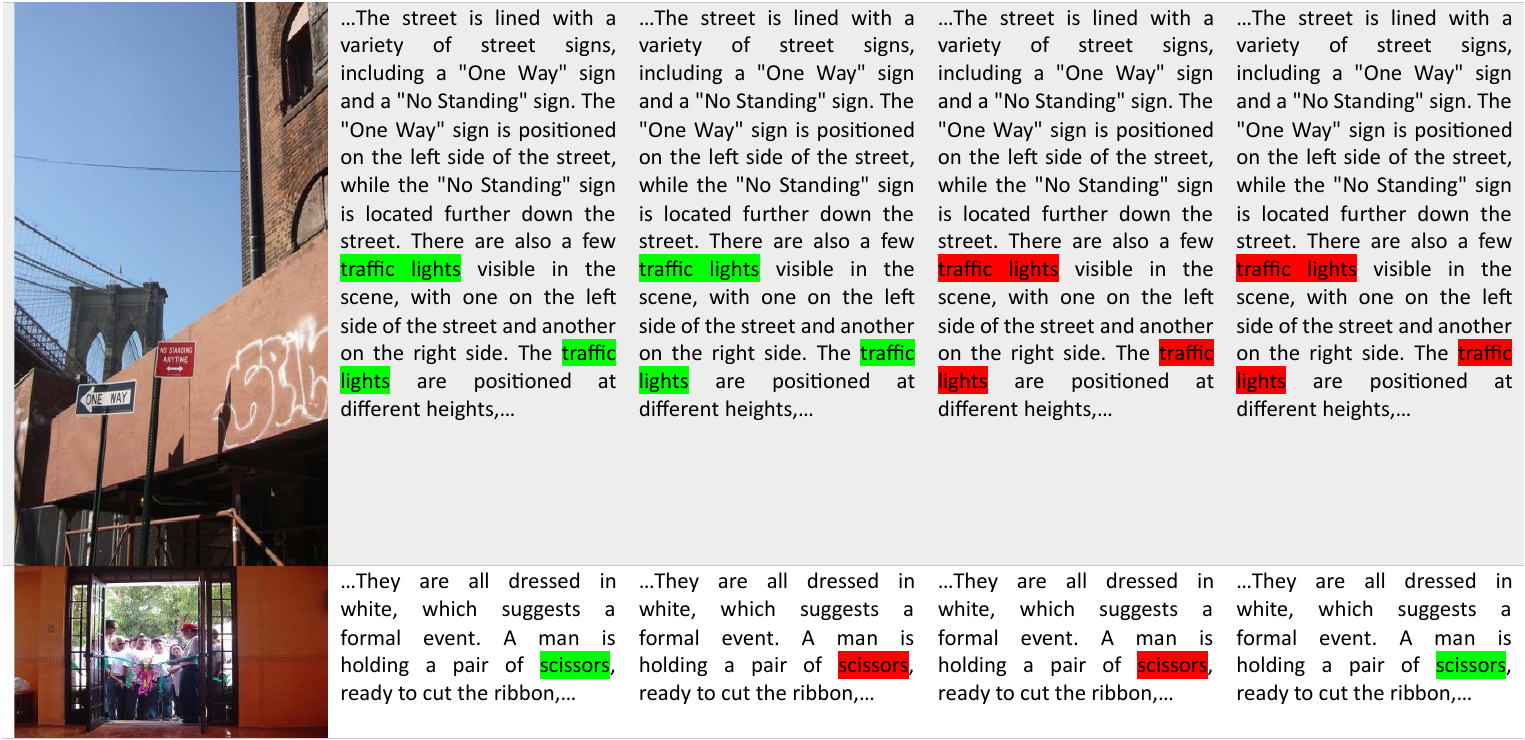}
    \end{minipage}
    \hfill
    \begin{minipage}{\textwidth}
        \includegraphics[width=\textwidth]{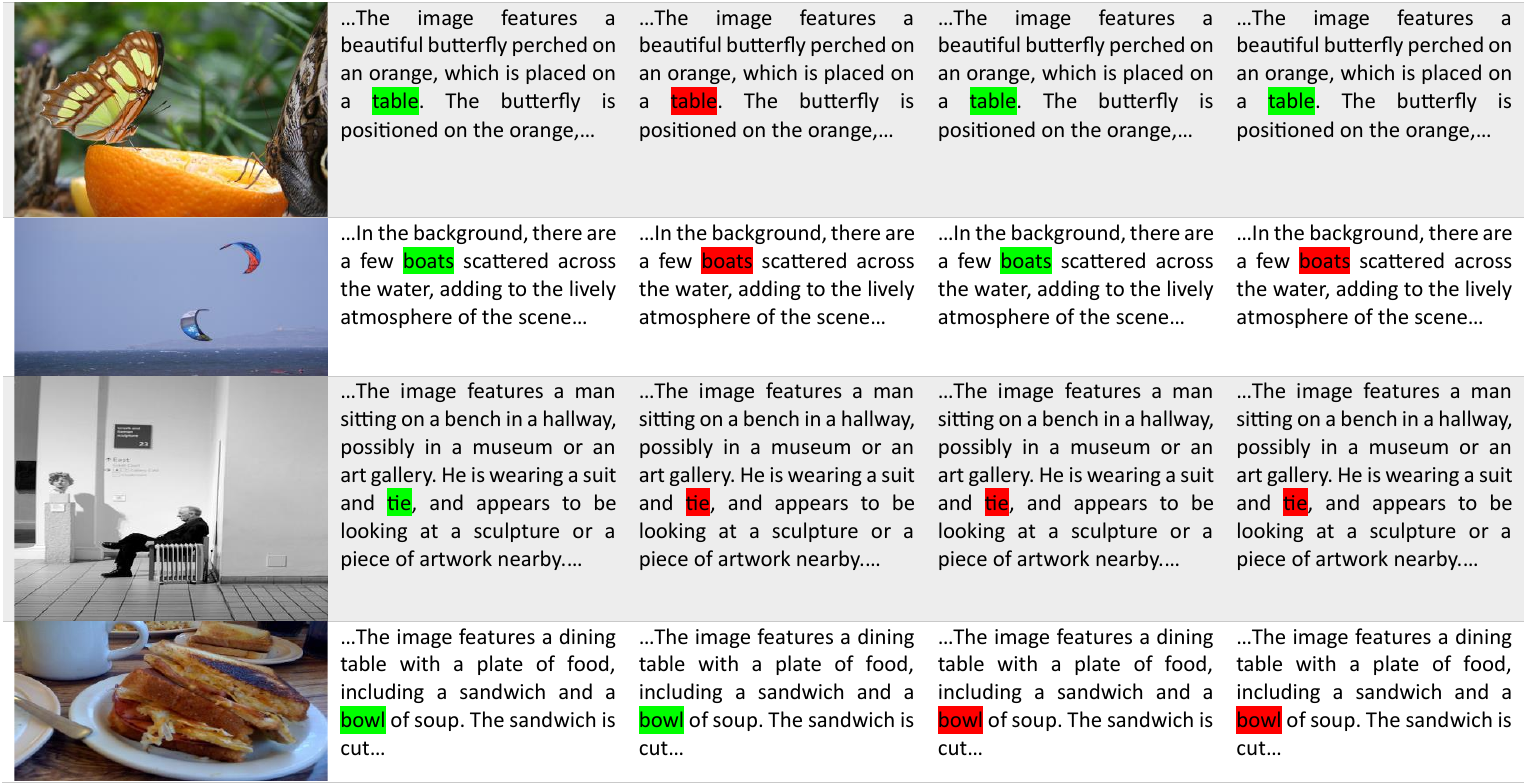}
    \end{minipage}
    \caption{More Qualitative Results.}
    \label{fig:q2}
\end{figure*}

\begin{figure*}[!htbp]
    \centering
    \begin{minipage}{\textwidth}
        \includegraphics[width=\textwidth]{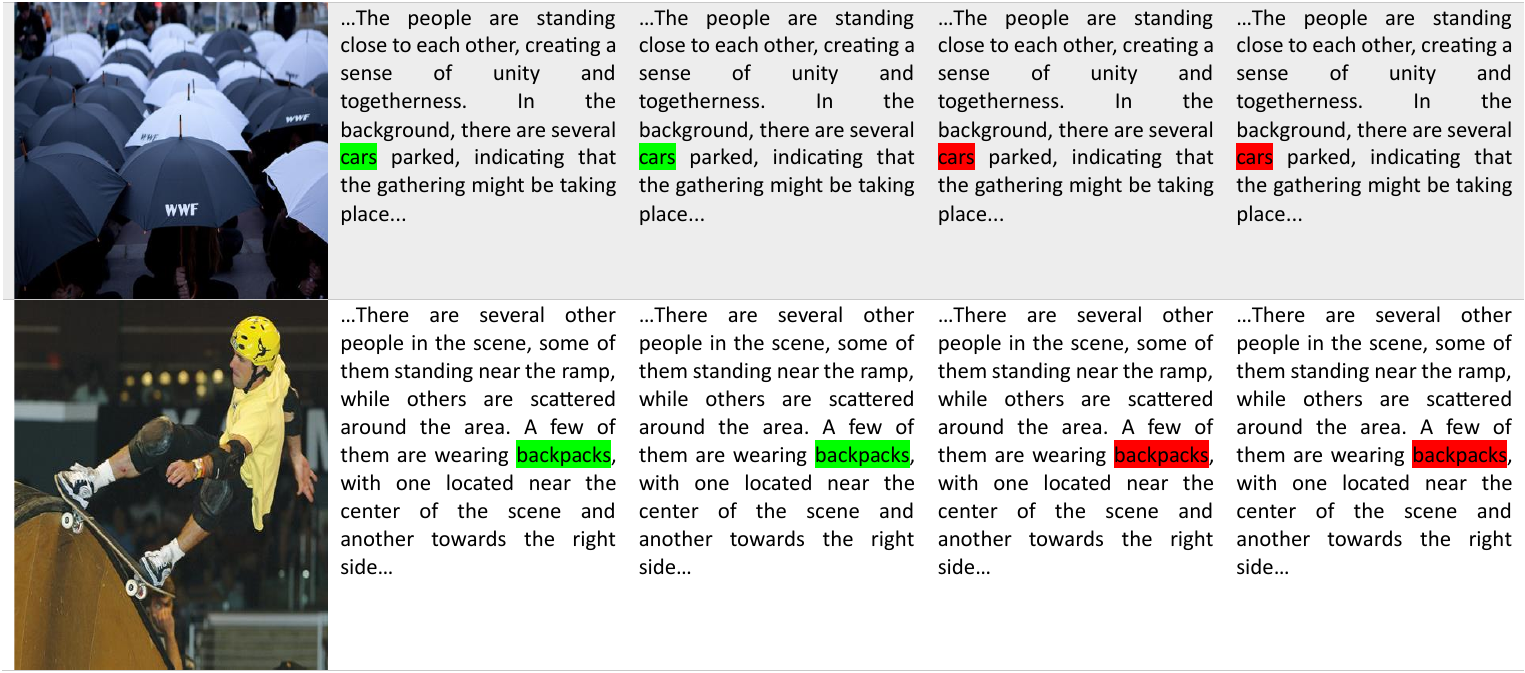}
    \end{minipage}
    \hfill
    \begin{minipage}{\textwidth}
        \includegraphics[width=\textwidth]{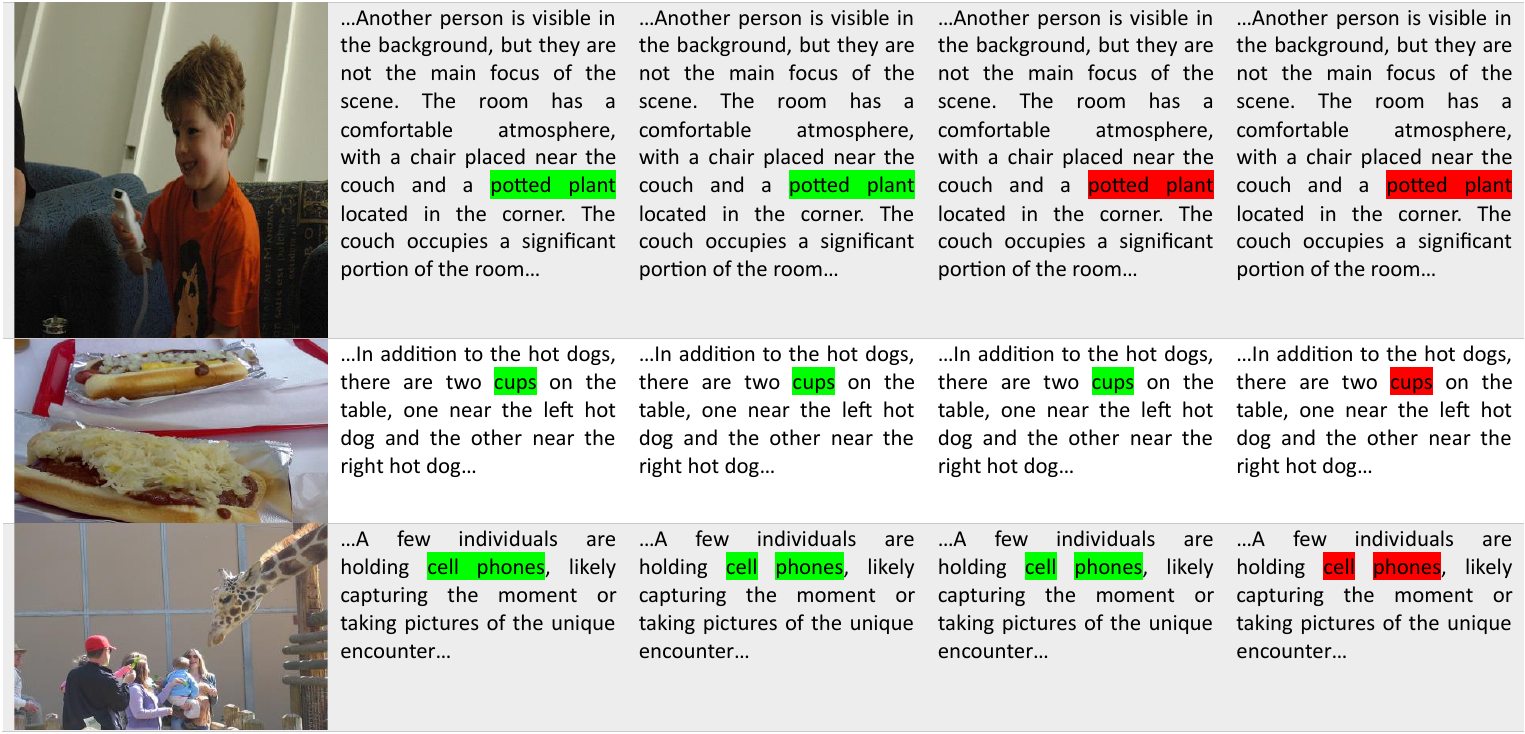}
    \end{minipage}
    \caption{More Qualitative Results.}
    \label{fig:q3}
\end{figure*}

\begin{figure*}[!htbp]
    \centering
    \begin{minipage}{\textwidth}
    \includegraphics[width=\textwidth]{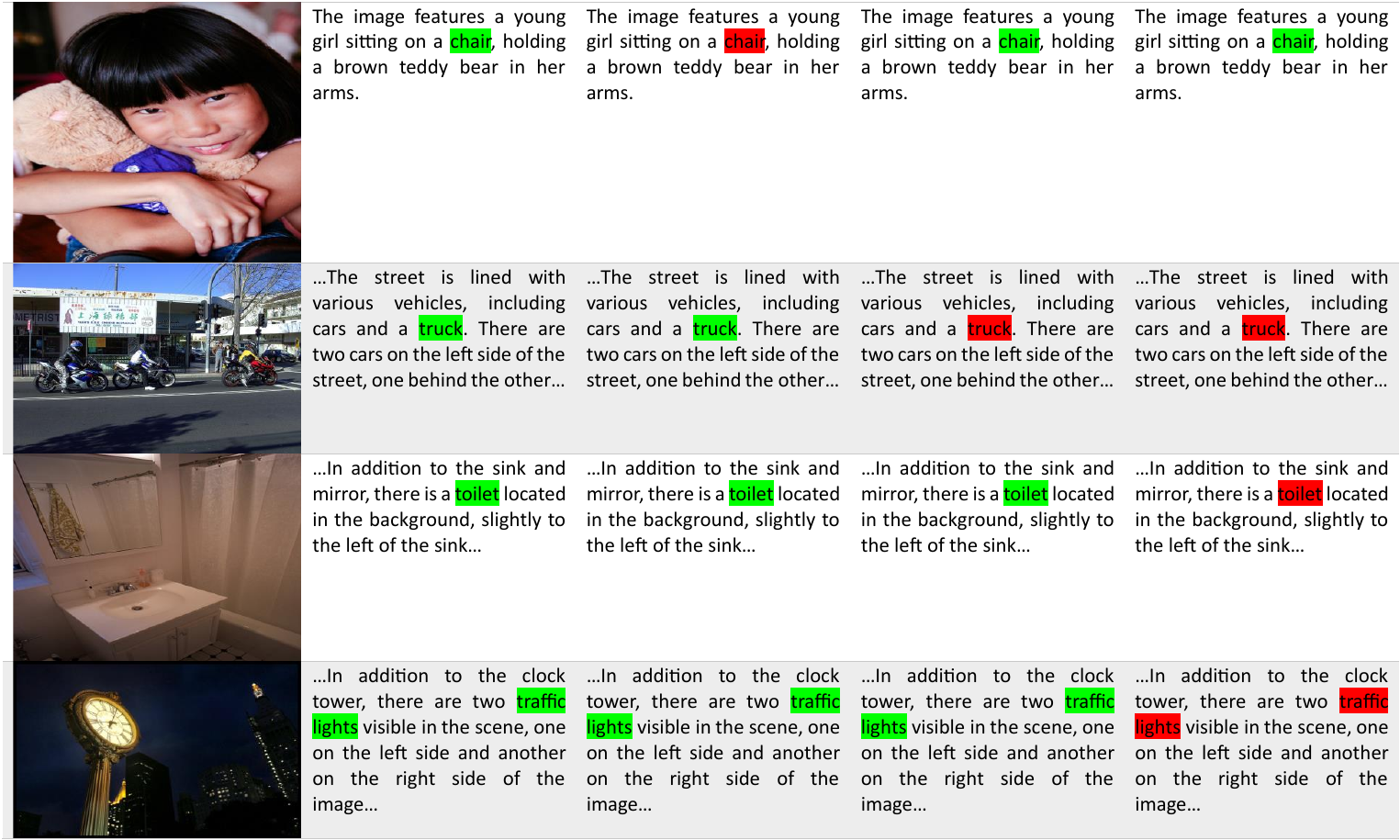}
    \end{minipage}
    \hfill
    \begin{minipage}{\textwidth}
        \includegraphics[width=\textwidth]{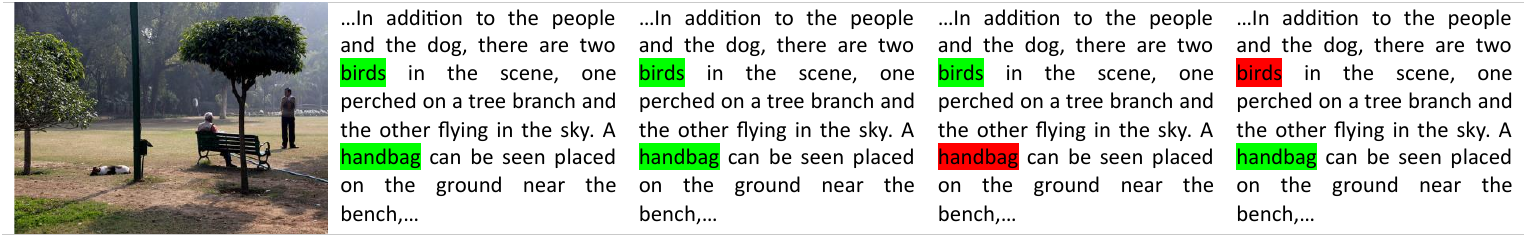}
    \end{minipage}
    \caption{More Qualitative Results.}
    \label{fig:q4}
\end{figure*}

We provide qualitative examples in Fig.~\ref{fig:q1},~\ref{fig:q2},~\ref{fig:q3},~\ref{fig:q4}.

\end{document}